\documentclass{article}
\usepackage{ijcai24}

\usepackage{times}
\usepackage{soul}
\usepackage{url}
\usepackage[hidelinks]{hyperref}
\usepackage[utf8]{inputenc}
\usepackage[small]{caption}
\usepackage{graphicx}
\usepackage{amsmath}
\usepackage{amsthm}
\usepackage{booktabs}
\usepackage{algorithm}
\usepackage{algorithmic}
\usepackage[switch]{lineno}

\usepackage{graphicx}

\usepackage{algorithm}
\usepackage{algorithmic}
\usepackage{amsmath}
\usepackage{amssymb}
\usepackage{natbib}

\usepackage{booktabs}

\usepackage[table]{xcolor}
\usepackage{tikz}
\usepackage{pgfplots}
\usepackage{tikzscale}
\usepackage{adjustbox}

\definecolor{lightblue}{RGB}{214,226,239}
\definecolor{customviolet}{RGB}{204,121,167}
\definecolor{lightviolet}{RGB}{222,213,255}
\definecolor{darkviolet}{RGB}{100,81,179}
\definecolor{darkerblue}{RGB}{60,19,223}
\definecolor{darkergreen}{RGB}{0,102,102}
\definecolor{darkgreen}{RGB}{0,100,0}

\definecolor{customred}{RGB}{213,94,0}
\definecolor{customblue}{RGB}{0,114,178}
\definecolor{customlightblue}{RGB}{86,180,233}
\definecolor{customorange}{RGB}{230,159,0}
\definecolor{customgreen}{RGB}{0,158,115}
\definecolor{custompink}{RGB}{204,121,167}
\definecolor{customyellow}{RGB}{230,219,0}

\newcommand{\Control}{\emph{Control}}
\newcommand{\Counterfactual}{\emph{Counterfactual}}
\newcommand{\Alterfactual}{\emph{Alterfactual}}
\newcommand{\Combination}{\emph{Combination}}
\title{Relevant Irrelevance: Generating Alterfactual Explanations for Image Classifiers}

\author{
Silvan Mertes$^1$
\and
Tobias Huber$^1$\and
Christina Karle$^1$\and
Katharina Weitz$^{1,2}$\and\\
Ruben Schlagowski$^1$\and
Cristina Conati$^3$\and
Elisabeth André$^1$\\
\affiliations
$^1$University of Augsburg, Germany\\
$^2$Fraunhofer HHI, Germany\\
$^3$University of British Columbia, Canada\\
\emails
\{silvan.mertes, tobias.huber, ruben.schlagowski, elisabeth.andre\}@uni-a.de,
katharina.weitz@hhi.fraunhofer.de,
conati@cs.ubc.ca
}

\begin{document}

\maketitle

\begin{abstract}
In this paper, we demonstrate the feasibility of alterfactual explanations for black box image classifiers.
Traditional explanation mechanisms from the field of Counterfactual Thinking are a widely-used paradigm for Explainable Artificial Intelligence (XAI), as they follow a natural way of reasoning that humans are familiar with. However, most common approaches from this field are based on communicating information about features or characteristics that are especially important for an AI's decision. 
However, to fully understand a decision, not only knowledge about relevant features is needed, but the awareness of irrelevant information also highly contributes to the creation of a user's mental model of an AI system. 
To this end, a novel approach for explaining AI systems called alterfactual explanations was recently proposed on a conceptual level. 
It is based on showing an alternative reality where irrelevant features of an AI's input are altered.
By doing so, the user directly sees which input data characteristics can change arbitrarily without influencing the AI's decision. 
In this paper, we show for the first time that it is possible to apply this idea to black box models based on neural networks.
To this end, we present a GAN-based approach to generate these alterfactual explanations for binary image classifiers.
Further, we present a user study that gives interesting insights on how alterfactual explanations can complement counterfactual explanations.
\end{abstract}

\section{Introduction}
With the steady advance of Artificial Intelligence (AI), and the resulting introduction of AI-based applications into everyday life, more and more people are being directly confronted with decisions made by AI algorithms \citep{stone2016one}. As the field of AI advances, so does the need to make such decisions explainable and transparent. The development and evaluation of \emph{Explainable AI} (XAI) methods is important not only to provide end users with explanations that increase acceptance and trust in AI-based methods, but also to empower researchers and developers with insights to improve their algorithms.

The need for XAI methods has prompted the research community to develop a large variety of different approaches to unravel the black boxes of AI models. A considerable part of these approaches is based on telling the user of the XAI system in various ways \emph{which} features of the input data are important for a decision (often called \emph{Feature Attribution}) \citep{arrieta2020explainable}. Other methods, which are close to human habits of explanation, are based on the paradigm of \emph{Counterfactual Thinking} \citep{miller2019explanation}. 
Procedures that follow this guiding principle try answering the question of \emph{What if...?} by showing an alternative reality and the corresponding decision of the AI.
Here, in contrast to feature attribution mechanisms, not only the importance of the various features is emphasized. Rather, it is conveyed, even if only indirectly, \emph{why} features are relevant.

\begin{figure}
 \includegraphics[width=0.48\textwidth]{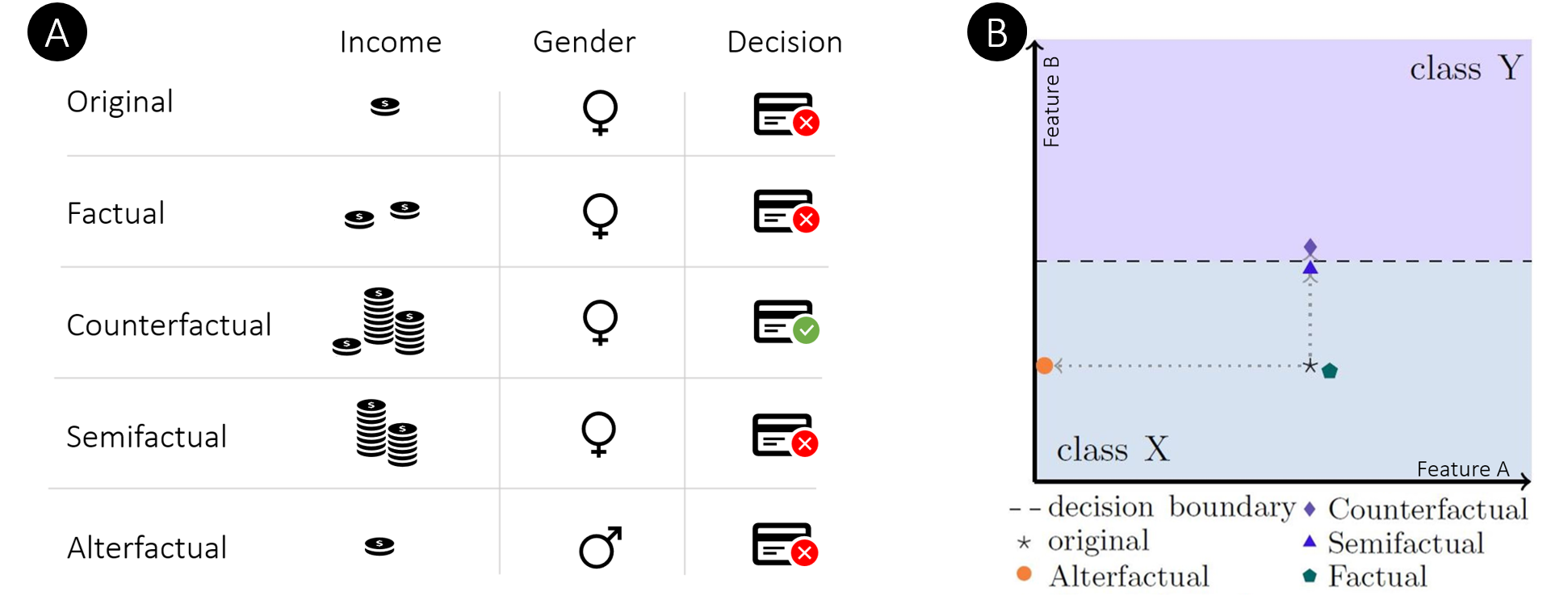}
 \caption{(A) Examples of a counterfactual and an alterfactual explanation. Input features to a fictional decision system to be explained are \emph{Income} and \emph{Gender}, whereas the former is relevant and the latter is irrelevant to the AI's decision on whether a credit is given or not.
%The diagram shows the original input which is to be explained below the decision boundary belonging to class X. A factual explanation could be the nearest neighbor, located anywhere around the original input. A semifactual explanation would be located in minimal distance directly next to the decision boundary, but still below it. A counterfactual explanation would be above it in the region of class Y, but barely so. An alterfactual explanation would move in parallel to the decision boundary, indicating which feature values would not modify the model's decision. Note that this diagram is highly simplified - normally, there are more than two features, the decision boundary is more complex, etc.
(B) Conceptual comparison of factual, counterfactual, semifactual,
and alterfactual explanations.} \label{fig:decision_boundary}
\end{figure}
Prominent examples of these explanatory mechanisms are \emph{Counterfactual Explanations} and \emph{Semifactual Explanations} \citep{kenny2020generating}.
Counterfactual explanations show a version of the input data that is altered just enough to change an AI's decision.  
By doing so, the user is shown not only \emph{which} features are relevant to the decision, but more importantly, \emph{how} they would need to be changed %in order to make the decision different. 
to result in a different decision of the AI.
Semifactual explanations follow a similar principle, but they modify the relevant features of the input data to an extent that the AI's decision does not change.

All of these methods have in common that they focus on the \emph{important} features.
However, we argue that awareness of irrelevant features can also contribute substantially to the complete understanding of a decision domain, as knowledge of the important features for the AI does not necessarily imply knowledge of the unimportant ones. 
%Therefor, the totality of the features per se would have to be known. Furthermore, common methods limit themselves to the \emph{most important} features, thus, the characteristics not treated can be either little important or irrelevant.

%For example, if we want to investigate whether an AI system is subject to some bias regarding its predictions, we often want to know explicitly whether a particular feature is completely irrelevant to a classifier.
For example, consider an AI system that assesses a person's creditworthiness based on various characteristics.
%, and we want to study that system regarding its fairness. 
If that system was completely fair, a counterfactual explanation might be of the form: \emph{If your income was higher, you would be creditworthy.}
However, this explanation does not exclude the possibility that your skin color also influenced the AI's decision.
It only shows that the income had a high impact on the AI. 
An explanation confined to the irrelevant features, on the other hand, might say \emph{No matter what your skin color is, the decision would not change.}
In this case, direct communication of irrelevant features ascertains fairness with regards to skin color.
Conventional counterfactual thinking explanation paradigms do not provide this information directly.
To address this issue, \citet{mertes2022alterfactual} recently conceptually introduced the explanatory paradigm of \emph{Alterfactual Explanations} that is meant to complement counterfactual explanations.
%We call explanations that follow this paradigm \emph{Alterfactual Explanations}. 
%We call explanations that follow this paradigm \emph{Alterfactual Explanations}.
This principle is based on showing the user of the XAI system an alternative reality that leads to the exact same decision of the AI, but where irrelevant features are altered. All relevant features of the input data, on the other hand, remain the same. 
As such, alterfactual explanations form the complement to counterfactual explanations - providing both explanation types should enable the user to grasp both relevant \emph{and} irrelevant features. 
Mertes et al. already showed the potential in the concept of raising user awareness about irrelevant features \citep{mertes2022alterfactual}. 
Nevertheless, due to the absence of an implementable solution, the researchers could only delve into the concept through the utilization of a fictional AI.

As such, in this work we introduce a GAN-based generation algorithm that is able to create both alterfactual and counterfactual explanations for image classifiers.\footnote{Our full implementation is open-source and available at \url{https://github.com/hcmlab/Alterfactuals}.}
As alterfactual explanations convey completely different information than common methods, we investigate whether the understanding that users have of the explained AI system is also formed in a different way, or can even be improved. 
Our results show that alterfactual explanations outperform counterfactual explanations with regards to local model understanding.

\section{Related Work}
As the approach presented in this paper can be counted to the class of XAI methods that work by inducing counterfactual thinking processes, it is important to gain an understanding of how common methods from this field work. Therefore, this section gives an overview on related explanation concepts.
Figure \ref{fig:decision_boundary}A illustrates the difference between those concepts using exemplary explanations for a fictional AI that decides if a person is creditworthy or not. We will use that scenario as a running example of how the different explanation paradigms would answer the question of \emph{Why does the AI say that I am not creditworthy?}.

\textbf{Factual Explanations} - \emph{There was another female person that also had rather little money, and she also did not get the credit.} - 
Factual explanations are the traditional way of explaining by example, and often provide a similar instance from the underlying data set (adapted or not) for the input data point that is to be explained \citep{keane2021twin}. Other approaches do not choose an instance from the dataset, but generate new ones \citep{guidotti2019factual}. The idea behind factual explanations is that similar data instances lead to similar decisions, and the awareness of those similarities leads to a better understanding of the model.
Further explanation mechanism that fall in this category are \emph{Prototypical Explanations} and \emph{Near Hits} \cite{kim2016examples, herchenbach2022explaining}.
%Thus, they aim to answer the question \emph{If the input would look more like this Factual Explanation, what would the decision be?}.

%\subsubsection{Contrastive Explanations}
%Contrastive explanations in the context of XAI answer the question of why a decision occurred relative to some other decision that could also have happened for a specific input \citep{stepin2021survey}. The decision that happened is often called \emph{fact}, while \emph{foil} refers to another possible decision it is being contrasted with \citep{miller2019explanation}. 
%Contrastive explanations are usually structured as \emph{Decision X as compared to decision Y occurred because features $f_1$ ... $f_n$ are present and features $f_1^\prime$ ... $f_n^\prime$ are absent} \citep{verma2020counterfactual}. 
%They, therefore, highlight the required minimally present and absent features to achieve the given decision.

\textbf{Counterfactual Explanations} - \emph{If you had that amount of money, you would get the credit.} - 
%Counterfactual explanations are often conflated with contrastive explanations, but do actually state which changes to a given input would be necessary to achieve a different decision by providing an example \citep{stepin2021survey}.
Counterfactual explanations are a common method humans naturally use when attempting to explain something and answer the question of \emph{Why not ...?} \citep{miller2019explanation, byrne2019counterfactuals}. 
In XAI, they do this by showing a modified version of an input to an AI system that results in a different decision than the original input.
Counterfactual explanations should be minimal, which means they should change as little as possible in the original input \citep{keane2021twin, miller2021contrastive}. %Many researchers have emphasized that counterfactual explanations should be actionable and feasible, i.e., should provide a user with an example that is achievable and realistic in real life \citep{barocas2020hidden, ustun2019actionable}. 
%How to achieve this is an ongoing research topic, with open questions for example whether feasibility can be achieved by adhering to data distributions found in the training set \citep{laugel2019dangers, mahajan2019preserving, keane2021if}. 
In certain scenarios, modern approaches for generating counterfactual explanations have shown significant advantages over feature attribution mechanisms (i.e., explanation approaches that highlight \emph{which} features are important for a decision) in terms of mental model creation and explanation satisfaction \citep{mertes2020ganterfactual}. 
\citet{wachter2017counterfactual} name multiple advantages of counterfactual explanations, such as being able to detect biases in a model, providing insight without attempting to explain the complicated inner state of the model, and often being efficient to compute. 
For counterfactual explanations, a multitude of works exist that, similar to how we do it for alterfactual explanations, use GANs to automatically generate explanations for image classifiers \cite{nemirovsky2022countergan, van2021conditional, khorram2022cycle, mertes2020ganterfactual}.
%Counterfactual explanations are a post-hoc, often model-agnostic, and local explanation method.

\textbf{Semifactual Explanations} - \emph{Even if you had that amount of money, you would still not get the credit.} -
Similar to counterfactual explanations, semifactual explanations are an explanation type humans commonly use. They follow the pattern of \emph{Even if X, still P.}, which means that even if the input was changed in a certain way, the prediction of the model would still not change to the foil \citep{mccloy2002semifactual}. In an XAI context, this means that an example, based on the original input, is provided that modifies the input in such a way that moves it toward the decision boundary of the model, but stops just before crossing it \citep{kenny2020generating}. Similar to counterfactual explanations, semifactual explanations can be used to guide a user’s future action, possibly in a way to deter them from moving toward the decision boundary \citep{keane2021twin}. 
%They are also post-hoc, usually model-agnostic, as well as local.

\section{Alterfactual Explanations} 
\emph{No matter what your gender is, the decicison would not change.} - 
The basic idea of alterfactual explanations investigated in this paper is to strengthen the user's understanding of an AI by showing irrelevant attributes of a predicted instance. 
Hereby, we define irrelevance as the property that the corresponding feature, regardless of its value, does not contribute in any way to the decision of the AI model.
When looking at models that are making decisions by mapping some sort of input data $x \in X$ to output data $y \in Y$, the so-called \emph{decision boundary} describes the region in $X$ which contains data points where the corresponding $y$ that is calculated by the model is ambiguous, i.e., lies just between different instances of $Y$. Thus, irrelevant features can be thought of as features that do not contribute to a data point's distance to the decision boundary.

However, information that is carried out by explanations should be communicated as clearly as possible. 
Alterfactual explanations inform about the \emph{irrelevance} of certain features - as such, it should be made clear that these features can take \emph{any} possible value.
%The information that is contained in an alterfactual explanation consists of the \emph{irrelevance} of certain features - as such, it should somehow be emphasized that these features can take \emph{any} possible value. 
If we would change the respective features only to a small amount, the irrelevance is not clearly demonstrated to the user.
%it is conceivable that the message of irrelevance would not become clear.
Therefore, an alterfactual explanation should change the affected features to the maximum amount possible. By doing so, they communicate that the feature, \emph{even if it is changed as much as it can change}, still does not influence the decision.
Thus, the definition of an alterfactual explanation is as follows:

\begin{quote}
Let $d: X \times X \rightarrow \mathbb{R}$ be a distance metric on the input space $X$. An \emph{alterfactual explanation} for a model $M$ is an altered version $a \in X$ of an original input data point $x \in X$, that maximizes the distance $d(x,a)$ whereas the distance to the decision boundary $B \subset X$ and the prediction of the model do not change: $d(x,B)=d(a,B)$ and $M(x)=M(a)$
\end{quote}

Thus, the main difference between an alterfactual explanation and a counterfactual or semifactual explanation  is that  while the latter methods alter features resulting in a decreased distance to the decision boundary, our alterfactual  method tries to keep that distance fixed. Further, while counterfactual and semifactual  methods try to keep the overall change to the original input minimal \citep{keane2021if, kenny2020generating}, alterfactual explanations do exactly the opposite, which is depicted in Figure \ref{fig:decision_boundary}B.

\section{Generating Alterfactual Explanations}

\begin{figure}[t]
\begin{adjustbox}{width=.34\textwidth,center}
\begin{tikzpicture}
\clip(-2,-6) rectangle (12,4);
\draw[->] (-1.2,0) -- (-0.8,0);
\draw[->] (-0.6,0) -- (10.95,0);

% layer
\fill[rounded corners,customblue] (-0.7, -2.7) rectangle (-0.3, 2.7) {}; % size 64
\fill[rounded corners,customlightblue] (0.2, -2.4) rectangle (0.6, 2.4) {}; % size 32
\fill[rounded corners,customlightblue] (1.1, -2.1) rectangle (1.5, 2.1) {}; % size 16
\fill[rounded corners,customlightblue] (2., -1.8) rectangle (2.4, 1.8) {}; % size 8
\fill[rounded corners,customlightblue] (2.9, -1.5) rectangle (3.3, 1.5) {}; % size 4
\fill[rounded corners,customlightblue] (3.8, -1.2) rectangle (4.2, 1.2) {}; % size 2
\fill[rounded corners,customyellow] (4.7, -0.9) rectangle (5.1, 0.9) {}; % size 1
\fill[rounded corners,customorange] (5.6, -1.2) rectangle (6., 1.2) {}; % size 2
\fill[rounded corners,customorange] (6.5, -1.5) rectangle (6.9, 1.5) {}; % size 4
\fill[rounded corners,customorange] (7.4, -1.8) rectangle (7.8, 1.8) {}; % size 8
\fill[rounded corners,customred] (8.3, -2.1) rectangle (8.7, 2.1) {}; % size 16
\fill[rounded corners,customred] (9.2, -2.4) rectangle (9.6, 2.4) {}; % size 32
\fill[rounded corners,customred] (10.1, -2.7) rectangle (10.5, 2.7) {}; % size 64
\fill[rounded corners,customgreen] (11, -3) rectangle (11.4, 3) {}; % size 128

% encoder - decoder separation
\draw[dashed] (5.35,-3.5) -- (5.35,3.3);

% names
\node[text width=3cm] at (0,0) {$x$};

\node[text width=3cm,rotate=90] at (-0.5,1.03) {\scriptsize $64\times64$};
\node[text width=3cm,rotate=90] at (0.4,1.03) {\scriptsize $32\times32$};
\node[text width=3cm,rotate=90] at (1.3,1.03) {\scriptsize $16\times16$};
\node[text width=3cm,rotate=90] at (2.2,1.15) {\scriptsize $8\times8$};
\node[text width=3cm,rotate=90] at (3.1,1.15) {\scriptsize $4\times4$};
\node[text width=3cm,rotate=90] at (4.,1.15) {\scriptsize $2\times2$};
\node[text width=3cm,rotate=90] at (4.9,1.15) {\scriptsize $1\times1$};
\node[text width=3cm,rotate=90] at (5.8,1.15) {\scriptsize $2\times2$};
\node[text width=3cm,rotate=90] at (6.7,1.15) {\scriptsize $4\times4$};
\node[text width=3cm,rotate=90] at (7.6,1.15) {\scriptsize $8\times8$};
\node[text width=3cm,rotate=90] at (8.5,1.03) {\scriptsize $16\times16$};
\node[text width=3cm,rotate=90] at (9.4,1.03) {\scriptsize $32\times32$};
\node[text width=3cm,rotate=90] at (10.3,1.03) {\scriptsize $64\times64$};
\node[text width=3cm,rotate=90] at (11.2,0.91) {\scriptsize $128\times128$};

% filter numbers
\node[text width=3cm,customgreen] at (0.85,-3) {\scriptsize $64$};
\node[text width=3cm,customgreen] at (1.7,-2.7) {\scriptsize $128$};
\node[text width=3cm,customgreen] at (2.6,-2.4) {\scriptsize $256$};
\node[text width=3cm,customgreen] at (3.5,-2.1) {\scriptsize $512$};
\node[text width=3cm,customgreen] at (4.4,-1.8) {\scriptsize $512$};
\node[text width=3cm,customgreen] at (5.3,-1.5) {\scriptsize $512$};
\node[text width=3cm,customgreen] at (6.2,-1.2) {\scriptsize $512$};

\node[text width=3cm,customgreen] at (7.1,-1.5) {\scriptsize $512$};
\node[text width=3cm,customgreen] at (8.,-1.8) {\scriptsize $512$};
\node[text width=3cm,customgreen] at (8.9,-2.1) {\scriptsize $512$};
\node[text width=3cm,customgreen] at (9.8,-2.4) {\scriptsize $256$};
\node[text width=3cm,customgreen] at (10.7,-2.7) {\scriptsize $128$};
\node[text width=3cm,customgreen] at (11.7,-3) {\scriptsize $64$};
\node[text width=3cm,customgreen] at (12.65,-3.3) {\scriptsize $1$};

% skip connections
% E1-D7
\draw (-0.05,0) -- (-0.05,-3.3); 
\draw (-0.05,-3.3) -- (10.75,-3.3);
\draw[->] (10.75,-3.3) -- (10.75,0);

% E2-D6
\draw (0.85,0) -- (0.85,-3);
\draw (0.85,-3) -- (9.85,-3);
\draw[->] (9.85,-3) -- (9.85,0);

% E3-D5
\draw (1.75,0) -- (1.75,-2.7);
\draw (1.75,-2.7) -- (8.95,-2.7);
\draw[->] (8.95,-2.7) -- (8.95,0);

% E4-D4
\draw (2.65, 0) -- (2.65, -2.4);
\draw (2.65,-2.4) -- (8.05,-2.4);
\draw[->] (8.05,-2.4) -- (8.05,0);

% E5-D3
\draw (3.55,0) -- (3.55,-2.1); 
\draw (3.55,-2.1) -- (7.15,-2.1);
\draw[->] (7.15,-2.1) -- (7.15,0); 

% E6-D2
\draw (4.5,0) -- (4.5,-1.8); 
\draw (4.5,-1.8) -- (6.25,-1.8);
\draw[->] (6.25,-1.8) -- (6.25,0);

% legend
\fill[rounded corners,customlightblue] (-1, -4.6) rectangle (-0.5, -4.3) {};
\fill[rounded corners,customorange] (-1, -5.) rectangle (-0.5, -4.7) {};
\fill[rounded corners,customred] (-1, -5.4) rectangle (-0.5, -5.1) {};

\fill[rounded corners,customblue] (7.5, -4.6) rectangle (8, -4.3) {};
\fill[rounded corners,customyellow] (7.5, -5.) rectangle (8, -4.7) {};
\fill[rounded corners,customgreen] (7.5, -5.4) rectangle (8, -5.1) {};
\node[text width=3cm,customgreen] at (9.15,-5.62) {$n$};

\node[text width=6cm] at(11.3,-4.45) {\scriptsize Conv2D + LeakyReLU};
\node[text width=6cm] at(11.3,-4.85) {\scriptsize Conv2D + ReLU};
\node[text width=6cm] at(11.3,-5.25) {\scriptsize Conv2DTrans + Tanh};
\node[text width=3cm] at(9.8,-5.59) {\scriptsize filter number};

\node[text width=6cm] at(2.8,-4.45) {\scriptsize Conv2D + BatchNorm + LeakyReLU};
\node[text width=8cm] at(3.8,-4.85) {\scriptsize Conv2DTrans + BatchNorm + Dropout + ReLU};
\node[text width=6cm] at(2.8,-5.25) {\scriptsize Conv2DTrans + BatchNorm + ReLU};

\end{tikzpicture}
\end{adjustbox}
 \caption{Architecture overview of the generator network.} \label{fig:architecture1}
\end{figure}

\begin{figure}[t]
\begin{adjustbox}{width=.35\textwidth,center}
\begin{tikzpicture}
\clip(-2,-1.5) rectangle (13.5,7);

% initial layer
\draw[rounded corners] (-0.05, 2.) rectangle (0.65, 4) {};
\draw[rounded corners] (1.45, 2.) rectangle (2.15, 4) {};

% main layer
\fill[rounded corners,customblue] (3.8, -0.5) rectangle (4.5, 4) {};
\fill[rounded corners,customlightblue] (5.3, 0) rectangle (6, 3.5) {};
\fill[rounded corners,customlightblue] (6.8, 0.5) rectangle (7.5, 3) {};
\fill[rounded corners,custompink] (8.3, 0.7) rectangle (9, 2.8) {};

% names
\node[text width=3cm] at (0,0.4) {$x$};
\node[text width=3cm] at (0,3) {$\hat{y}$};
\node[text width=3cm] at (4.4,1.65) {$+$};

\node[text width=3cm,rotate=90] at (0.3,3.52) {\textit{Embedding}};
\node[text width=3cm,rotate=90] at (1.8,3.57) {\textit{Upsample}};

\node[text width=3cm,rotate=90] at (4.15,2.6) {$64\times64$};
\node[text width=3cm,rotate=90] at (5.65,2.65) {$32\times32$};
\node[text width=3cm,rotate=90] at (7.15,2.7) {$31\times31$};
\node[text width=3cm,rotate=90] at (8.65,2.7) {$30\times30$};

\node[text width=3cm,customgreen] at (5.5,-0.9) {\footnotesize $64$};
\node[text width=3cm,customgreen] at (6.95,-0.4) {\footnotesize $128$};
\node[text width=3cm,customgreen] at (8.45,0.1) {\footnotesize $256$};
\node[text width=3cm,customgreen] at (10.1,0.25) {\footnotesize $1$};

\draw[->] (-1.1,3) -- (-0.3,3); 
\draw[->] (0.8,3) -- (1.2,3); 
\draw[->] (2.3,3) -- (2.85,1.9); 
\draw[->] (-1.1,0.45) -- (2.8,1.5); 

\draw[->] (3.3,1.65) -- (3.6,1.65); 

\draw[->] (4.7,1.65) -- (5.1,1.65); 
\draw[->] (6.2,1.65) -- (6.6,1.65); 
\draw[->] (7.7,1.65) -- (8.1,1.65);

% legend
\fill[rounded corners,customblue] (6, 6.3) rectangle (6.5, 6.6) {};
\fill[rounded corners,customlightblue] (6, 5.9) rectangle (6.5, 6.2) {};
\fill[rounded corners,custompink] (6, 5.5) rectangle (6.5, 5.8) {};

\node[text width=3cm,customgreen] at(7.65,5.24) {$n$};

\node[text width=6cm] at(9.99,6.45) {\footnotesize Conv2D + LeakyReLU};
\node[text width=6cm] at(9.99,6.05) {\footnotesize Conv2D + BatchNorm + LeakyReLU};
\node[text width=4cm] at(8.99,5.65) {\footnotesize Conv2D + Sigmoid};
\node[text width=3cm] at(8.5,5.25) {\footnotesize filter number};

\end{tikzpicture}
\end{adjustbox}
 \caption{Architecture overview of the discriminator network.} \label{fig:architecture2}
\end{figure}

As we argue that alterfactual and counterfactual explanations convey different information, we designed a generative approach that is capable of creating both types of explanations in order to explain an image classifier.
For both, a set of requirements arises that needs to be reflected in the objectives of our explanation generation approach.
\begin{enumerate}
    \item The generated explanations should have high quality and look realistic.
    \item The resulting explanation should be either classified as the same class as the original input (for alterfactual explanations), or as the opposite class (for counterfactual explanations).
    \item For alterfactual explanations, the output image should change as much as possible, while for counterfactual explanations, it should change as little as possible.
    \item For alterfactual explanations, only irrelevant features should change, i.e., the distance to the decision boundary should be maintained.
\end{enumerate}
To address these objectives, different loss components (see next sections) were used to steer a GAN-based architecture to generate the desired explanations. A GAN-based approach was chosen as similar concepts have successfully been applied to the task of counterfactual explanation generation in various existing works \citep{olson2021, huber2023ganterfactual-rl, nemirovsky2022countergan, zhao2020fast, mertes2020ganterfactual}.
In order to allow for a more focused and comprehensive user study design, in this work, we focus on explaining a binary image classifier. However, although our specific generation architecture is designed for a binary classification problem, it would theoretically be possible to apply it to non-binary tasks by training separate models for each class vs. the union over all other classes.
%However, it should be noted that our architecture can be adopted to a multi-class problem without much effort.
% To address those objectives,  we use different loss components to steer a GAN-based architecture to generate the desired explanations. 
A schematic overview of our architecture can be seen in Figures \ref{fig:architecture1} and \ref{fig:architecture2}. For a more detailed description, we refer to the appendix. 
%For this work, we focus on explaining binary classifiers $C$, that has to be 

\subsection{Adversarial Component}
To address the first objective, an adversarial setting is used. 
Here, a generator network $G$ is trained to take an original image $x$ and a random noise vector $z$ and transforms them into the respective explanation $\hat{x}$. 
As such, a mapping $\{x,z\} \rightarrow \hat{x}$ is learned by the generator.
A discriminator network $D$ is trained to identify the generated images as \emph{fake} images in an adversarial manner.

Additionally, to partly target the second objective, we feed a target class label $\hat{y} \in \{0, 1\}$ to the discriminator. By doing so, the discriminator learns not only to assess if the produced images are real or fake, but also has the capability to decide if an explanation fits the data distribution of the class it is supposed to belong to. A somewhat similar idea was put forth by \citet{sharmanska2020contrastive} within the context of fairness and yielded promising results there.
During training, the discriminator is alternately fed with real and fake data. For real data, the target class label $\hat{y}$ reflects the class that the classifier to be explained assigns to the respective image $x$. For the generated explanations, the target class label $\hat{y}$ reflects either the class that was assigned to the original image $x$ (for alterfactual explanations), or the opposite class (for counterfactual explanations).

By letting the generator and discriminator compete against each other during training, it is enforced that the resulting images look realistic and resemble the data distribution of the respective target classes.
The objective function for the adversarial setting is formulated as follows:
%With $p_{data(\hat{x})}$ referring to the data distribution of the target domain, the objective is:
\begin{multline}
 \mathcal{L}_{adversarial} = \mathbb{E}_{x \sim p_{\text{data}}(x)} \left[ \log D(x, \hat{y}) \right] + \\
 \mathbb{E}_{x \sim p_{\text{data}}(x), z \sim p_{\text{noise}}(z)} \left[ \log(1 - D(G(x, z),\hat{y})) \right]
\end{multline}

\subsection{Including Classifier Information}
The second objective is further addressed by incorporating the decisions of the classifier to be explained into the generator's loss function.
%Ideally, it should also be possible to encourage the generator to output images with a similar distance to the decision boundary of the classifier (for Alterfactuals). 
%For Counterfactuals, this is not a requirement, but it did not empirically impact their performance if a loss function encouraging this was used - the effect most likely mitigated by the other components of the overall loss function. 

%For Alterfactuals,  is the simple difference between the classifier output for the original $x \in X$ and the generated image $\hat{x}$.

Let $C: X \rightarrow \left[0,1\right]$ be a binary classifier with threshold $0.5$.
% For a binary classifier $C: X \rightarrow \left[0,1\right]$,
We define the classification target $\tilde{C}(x)$ as $ \tilde{C}(x) := C(x) $ for \text{alterfactual explanations} and $\tilde{C}(x) := 1 - C(x)$ for \text{counterfactual explanations}.
To measure the error between the actual classification of the generated explanation and the target classification, we used Binary Crossentropy (BCE) to define a classification loss $\mathcal{L}_{C}$:
\begin{multline}
    \mathcal{L}_{C} = \mathbb{E}_{x, \hat{x}\sim p_{data}(x, \hat{x})}\lbrack \tilde{C}(x) \cdot \log C(\hat{x}) \\
    + (1 - \tilde{C}(x)) \cdot \log(1 - C(\hat{x}))\rbrack
\end{multline}

\subsection{SSIM Component}
The third objective was addressed by including a similarity component into the loss function. 
Explanations are meant for humans. Therefore, using the Structural Similarity Index (SSIM) seemed to be an appropriate choice to measure image similarity for our approach, as it correlates with how humans are perceiving similarity in images \citep{wang2004image}.
The parameters for SSIM were chosen as recommended by \citet{wu2019privacy}.

%chose $\alpha = \beta = \gamma = 1$, $K_{1} = 0.01$, $K_{2} = 0.03$, $C_{3} = C_{2} / 2$, and $L = 1$, as recommended by \cite{WU ET AL}.

% The SSIM is based on the mean and standard deviation of images:
% \begin{equation}
%     \mu_{x} = \frac{1}{N} \sum_{i=1}^{N}x_{i}
%     \quad\mathrm{and}\quad
%     \sigma_{x} = (\frac{1}{N - 1} \sum_{i=1}^{N}(x_{i} - \mu_{x})^{2})^{\frac{1}{2}}
% \end{equation}

% where $N$ is the number of images to be compared - in our case the original image and the explanation ($N=2$).
% Mean and standard are then used to calculate three sub-components of this metric: luminance, contrast, and structural similarity. 
% Luminance similarity is defined as:
% \begin{equation}
%     l(x, y) = \frac{2\mu_{x}\mu_{y} + C_{1}}{\mu_{x}^{2} + \mu_{y}^{2} + C_{1}}
%     \quad\mathrm{with}\quad
%     C_{1} = (K_{1}L)^{2}.
% \end{equation}

% Contrast similarity is measured by:
% \begin{equation}
%     \frac{2\sigma_{x}\sigma_{y} + C_{2}}{\sigma_{x}^{2} + \sigma_{y}^{2} + C_{2}}
%     \quad\mathrm{with}\quad
%     C_{2} = (K_{2}L)^{2}
% \end{equation}

% Structural similarity is defined as:
% \begin{multline}
%     s(x, y) = \frac{\sigma_{xy} + C_{3}}{\sigma_{x}\sigma_{y} + C_{3}}  \\
%     \quad\mathrm{with}\quad
%     \sigma_{xy} = \frac{1}{N - 1} \sum_{i=1}^{N}(x_{i} - \mu_{x})(y_{i} - \mu_{y})
% \end{multline}

% The three components are then combined to form the overall metric:
% \begin{equation}
%     SSIM(x, y) = \left[l(x, y)\right]^{\alpha} \cdot \left[c(x, y)\right]^{\beta} \cdot \left[s(x, y)\right]^{\gamma}
% \end{equation}

As alterfactual explanations should change irrelevant features \emph{as much as possible}, while counterfactual explanations should be \emph{as close as possible} to the original image, the learning objective differs for both (low similarity for alterfactual explanations, high similarity for counterfactual explanations).
With $\left[0,1\right]$ as the range of SSIM, we designed the loss function as follows:
 \begin{equation}
     \mathcal{L}_{sim} = \begin{cases}
\mathbb{E}_{x, \hat{x}\sim p_{data}(x, \hat{x})}\left[SSIM(x, \hat{x})\right]  \text{\hspace{\fontdimen2\font}\hspace{\fontdimen2\font}\hspace{\fontdimen2\font}\hspace{\fontdimen2\font}\hspace{\fontdimen2\font}\hspace{\fontdimen2\font}\hspace{\fontdimen2\font}Alterfactual}\\
\mathbb{E}_{x, \hat{x}\sim p_{data}(x, \hat{x})}\left[1 - SSIM(x, \hat{x})\right]  \text{Counterfactual} 
\end{cases}
 \end{equation}
 
\subsection{Feature Relevance Component}
The fourth objective, i.e., forcing the network to only modify irrelevant features when generating alterfactual explanations, was addressed by using an auxiliary Support Vector Machine (SVM) classifier. Note that this loss is only applied when generating alterfactual explanations, not when generating counterfactual explanations.
\citet{li2018decision} and \citet{elsayed2018large} have shown theoretically and empirically that the last weight layer of a Neural Network converges to an SVM trained on the data transformed up to this layer if certain restrictions are met (e.g., the last two layers of the network have to be fully connected).
An SVM's decision boundary can be calculated directly - unlike the one of a Neural Network \citep{jiang2018predicting}.
As such, we use an SVM which was trained to predict the classifier's decision based on the activations of the classifier's penultimate layer as a way to approximate the classifier's decision boundary - if the generated alterfactual explanation has moved closer to the SVM's separating hyperplane, relevant features were most likely modified. Although an unchanged decision boundary distance does not necessarily guarantee that no relevant features were modified, in our experiments, it was a good indicator.

The distance of $x$ to the SVM's separating hyperplane $f$ was defined as follows, with $w$ as the SVM's weight vector:
\begin{equation} \label{formula:SVM_DB}
    \text{\textit{SVM}}(x) = \left\lvert\frac{f(x)}{||w||}\right\lvert
\end{equation}

%The difference between the original image and the generated Alterfactual is calculated as their absolute difference in distance to the separating hyperplane:
The SVM loss is defined by the absolute difference in distance to the separating hyperplane between the original image and the generated alterfactual explanation:
\begin{equation}
    \mathcal{L}_{\text{\textit{SVM}}} = \mathbb{E}_{x\sim p_{data}(x), z\sim p_{\text{noise}}(z)}\left[|\text{\textit{SVM}}(x) - \text{\textit{SVM}}(\hat{x})|\right]
\end{equation}

The final loss function is a summation of all the four loss components introduced above.

\section{Evaluation Scenario}
\label{sec:evaluation_scenario}
\begin{figure}
    \centering
    % \captionsetup{justification=raggedright,margin=.5cm}
    \includegraphics[width=1\linewidth]{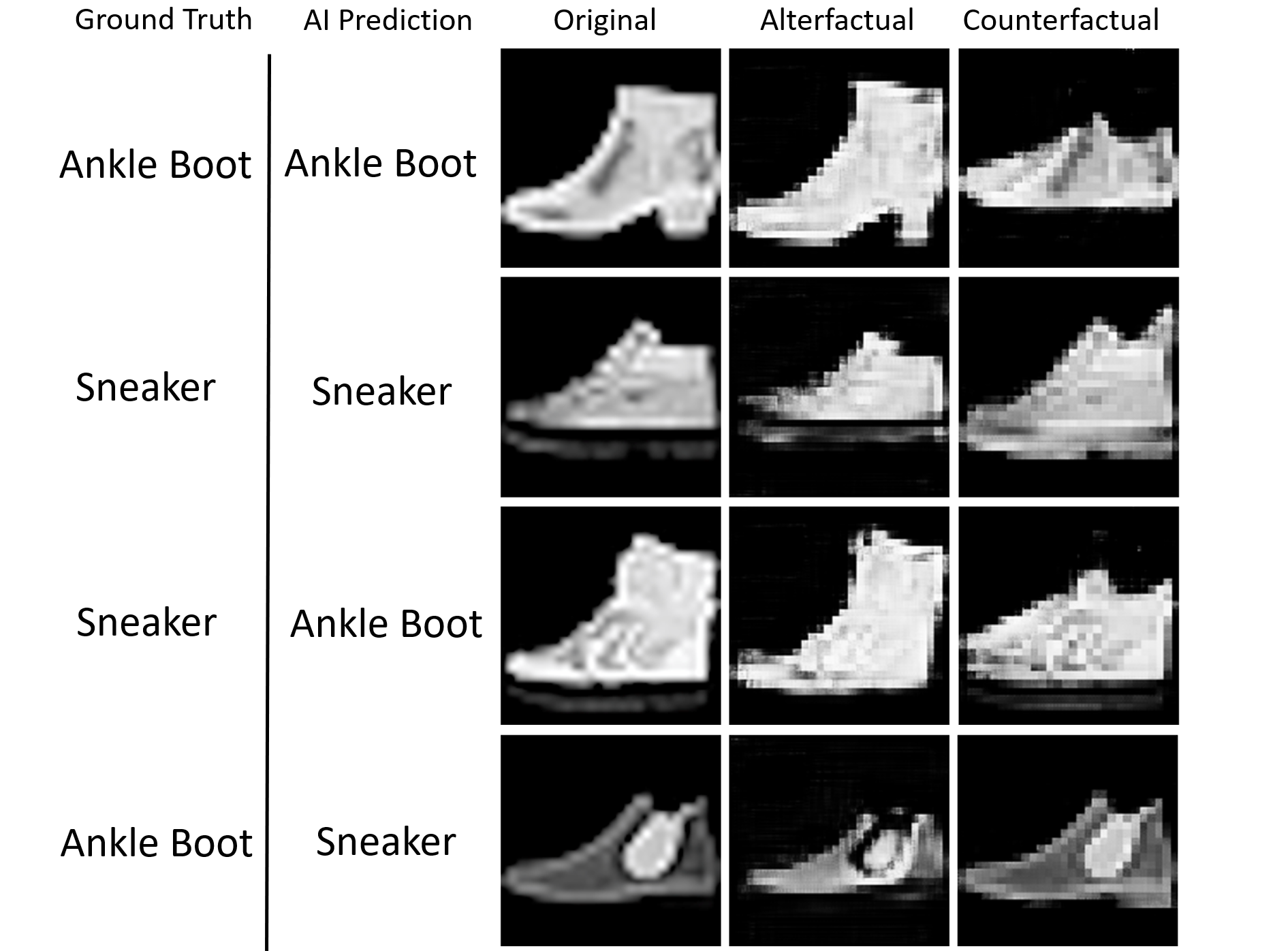}
     \caption{Example outputs of our system. It can be seen that alterfactual explanations change features that are irrelevant to the classifier, e.g., the color of the shoes or the width of the boot shaft, while counterfactual explanations change relevant features like the presence or absence of a boot shaft. From top to bottom the original images are a correctly classified ankle boot and sneaker, followed by two inputs incorrectly classified as ankle boot and sneaker.}
    \label{fig:example_outputs}
\end{figure}

To assess the performance of our approach, we applied it to the Fashion-MNIST data
set \citep{xiao2017fashion}.
That data set contains 7,000 gray scale images for each of its ten categories of clothes, such as `ankle boots' or `pullover', splitted into \emph{train} (6,000 images per class) and \emph{test} (1,000 images per class) sets.
%The two classes chosen for this data set, `ankle boots' and `sneakers', were selected due to being somewhat similar in order not to oversimplify the classification task, while still distinct enough to be able to visually assess whether the generated explanations seem plausible.
The two classes we chose, `ankle boots' and `sneakers', were selected due to being somewhat similar in order not to oversimplify the classification task while still being distinct enough to be able to visually assess whether the generated explanations are clear.
%As classifier to be explained, we trained a rather simple four-layer convolutional neural network, achieving an accuracy of $96.7\%$ after 40 epochs of training.
To create the classifier to be explained, we trained a relatively simple four-layer convolutional neural network, achieving an accuracy of $96.7\%$ after 40 training epochs.
The exact architecture and training configuration can be found in the appendix.

Our explanation generation architecture was trained for 14 epochs, until visually no further improvement could be observed. For alterfactual explanations, we reached a validity (i.e., which portion of the explanations are classified as the correct target class by the classifier) of $96.20\%$ and an average SSIM of $0.32$ (here, lower is better), whereas the counterfactual explanations reached a validity of $87.70\%$ and an average SSIM of $0.90$ (here, higher is better). For more details refer to the appendix.
Exemplary generated explanations are shown in Figure \ref{fig:example_outputs}.
Note that, in order to verify if our alterfactual generation approach is applicable on a wider range of datasets, we additionally trained our approach on three other datasets: MNIST \citep{lecun1998mnist}, MaskedFace-Net \citep{cabani2021maskedface}, and a gray scale version of MaskedFace-Net. 
To further demonstrate that our approach can be adapted to be more model-agnostic and work without access to intermediate layers, we omitted the Feature Relevance component for those experiments.
%To demonstrate not only the applicability to different datasets, but also to show that our approach can simply be adapted to be model-agnostic, we omitted the Feature Relevance component for those experiments. 
Training details, example outputs and computational results for those experiments can be found in the appendix.

\section{User Study}

\subsection{Research Question and Hypotheses}
We conducted a user study to validate whether the counterfactual and alterfactual explanations generated by our approach help human users to form correct model understanding of an AI system. 
Therefore, we only used results from the model trained on the Fashon-MNIST classifier in order to not overwhelm participants.
To be able to compare our findings to existing work, we designed our study similar to \citet{mertes2022alterfactual}. Our hypotheses are as follows: 

% \begin{itemize}
%     \item A combination of alterfactual and counterfactual explanations is the most effective way to enable a good model understanding.
%     \item A combination of alterfactual and counterfactual explanations is most satisfying for users.
%     \item Both alterfactual and counterfactual explanations are more effective in enabling model understanding than no explanations.
%     \item Alterfactual explanations are more effective to identify irrelevant features while counterfactual explanations should help more with identifying relevant features.
% \end{itemize}
% \begin{itemize}
    1) Alterfactual and counterfactual explanations, as well as the combination of both, are more effective in enabling model understanding than no explanations.
    
    2) There is a difference in model understanding and explanation satisfaction between alterfactual and counterfactual explanations, but we did not anticipate a specific direction since we see them as complementary concepts.
    
    3) Compared to the individual explanations, a combination of alterfactual and counterfactual explanations is a more effective way to enable a good model understanding and is more satisfying for users. 
    % \item A combination of alterfactual and counterfactual explanations is most satisfying for users.
    %\item Between alterfactual and counterfactual explanations we investigate non-directional hypotheses regarding model understanding and explanation satisfaction.
    
    4) There is a difference between conditions regarding the understanding of relevant and irrelevant features, where alterfactual explanations are more effective to identify irrelevant features while counterfactual explanations should help more with identifying relevant features.
%\end{itemize}
%We hypothesized that 
%Moreover, we thought that both alterfactual and counterfactual explanations are still more effective in enabling model understanding than no explanations. 
%Moreover, we thought that both alterfactual and counterfactual explanations on their own are still more effective in terms of enabling model understanding than no explanations. 
%Since we see alterfactual and counterfactual explanations as complementary concepts, we did not have a general hypothesis between the two explanations. 
%Further, we thought that alterfactual explanations would be more effective to identify irrelevant features while counterfactual explanations should help more with identifying relevant features.

\subsection{Methodology}
\paragraph{Conditions and Explanation Presentation} 
We used a between-groups design with four conditions.
Participants in the \Control{} condition were presented only with the original input images to the AI. No explanation was shown.
In the \Alterfactual{} and \Counterfactual{} conditions, participants were presented with the original input images and either alterfactual or counterfactual explanations.
%The explanations were presented through a slider under the original image. When the slider is moved, the image gets linearly interpolated into the counterfactual or alterfactual explanation.
In the \Combination{} condition, participants were presented with the original input images as well as both the alterfactual and the counterfactual explanations.
%In this case the slider could be moved in two directions to morph the original image into an counterfactual or alterfactual explanation (Figure \ref{fig:explanation_presentation}). The order of the sides was randomized.

\iffalse
Figure \ref{fig:explanation_presentation} shows how the explanations were presented in each condition.

\begin{figure}
    \centering
    % \captionsetup{justification=raggedright,margin=.5cm}
    \includegraphics[width=.5\linewidth]{figures/shoe_with_slider2.png}
     \caption{An example of how the explanations were presented in the \Combination{} condition during the user study. By moving the slider to either side, the image is linearly interpolated into the counterfactual or alterfactual explanation. The order of the sides was randomized. In the \Alterfactual{} and \Counterfactual{} conditions, only one side of the slider was present and in the \Control{} condition, there was no slider.}
    \label{fig:explanation_presentation}
\end{figure}
\fi

\paragraph{Procedure}
The whole study was built using the \emph{oTree} framework by \citet{chen2016otree}.
After answering questions about their demographic background, participants were given some general information about the data and their task during the prediction task. 
For the classifier, they were only told that an AI was trained to distinguish between ankle boots and sneakers.
Two example images for each class (ankle boots and sneakers) were shown and some shoe specific terminology (e.g., "shaft") was introduced in order to make sure that participants have a common understanding of the terms they are asked about later on.
Following this information, the participants were given an example input image for each class together with the classifier's prediction for this input image. 
In the explanation conditions, the participants were introduced to their corresponding explanation types (counterfactuals, alterfactuals or a combination) and could explore the explanations for those two images.
After that, each participant answered a quiz about the information that was given up to that point, to make sure that they understood everything correctly. 
Subsequently, the study itself started. 
It was divided into three parts: For assessing the participants' understanding of the classifier, we used \emph{(i)} a prediction task for assessing the local understanding, i.e., to assess if the participants understand why the AI makes a \emph{specific} decision, and \emph{(ii)} a questionnaire about the relevance of certain features for assessing the global understanding, i.e., to assess if the participants understand how the AI works \emph{overall}. To assess the participants' explanation satisfaction, we used \emph{(iii)} an explanation satisfaction questionnaire. The three phases of the experiment are described below.

\paragraph{Local Model Understanding: Prediction Task}
To measure the local understanding of the classifier, we used a prediction task, which assesses the participants' ability to anticipate the AI classifier's decisions \citep{hoffman2018metrics}.
Eight examples were shown, covering all possible classification outcomes (two correctly classified images for both sneakers and ankle boots, and two incorrectly classified images for both) to avoid bias.
Figure \ref{fig:example_outputs} shows four of the images from the study.
The example images were chosen randomly but we made sure that the alterfactual and counterfactual explanations generated by our model for those images were valid (i.e., the classifier predicted the same class as for the original image when fed with the alterfactual explanation, and the opposite class when fed with the counterfactual explanation).
Participants had to predict the classifier's decision for each example image. 
Participants were additionally asked about their own opinion on which class the original shoe image belonged to. The answers to that particular question were not further analyzed - it was only added to help the participants distinguish between their own opinion and their understanding of the classifier.
%To help the participants to distinguish between their own opinion and their understanding of the classifier, they were additionally asked about their own opinion on which class the original shoe image belonged to.
%For both predictions they rated their confidence on a 7-point Likert scale.
After predicting an example, they were told the correct label and the AI classifier's decision before moving on to the next example.
The order of the examples was randomized.
% Participants in the "No Explanation" condition lacked explanations and relied solely on the original input data. 
% In the epxlanation conditions, the participants could reveal the explanations by using a slider under the original image to morph the image into a coutnerfactual or alterfactual image/explantion depending on the condition. 

\paragraph{Global Model Understanding: Feature Relevance}
While the Prediction Task can be seen as \emph{local} measurement of the users' understanding of the model in specific instances, we also wanted to investigate whether participants understood the \emph{global} relevance of different features.
To this end, we looked at two  features that were relevant for our classifier ("presence/absence of a boot shaft" and "presence/absence of an elevated heel") as well as two features that were irrelevant for our classifier ("boot shaft width" and "the shoe's color and pattern on the surface area"). 
These features were chosen based on the authors' experience from training the classifier and a-priori explorations with the Feature Attribution explanation mechanisms LIME \citep{ribeiro2016LIME} and SHAP \citep{lundberg2017SHAP}. 
Note that, although the classifier is still a black box and there is no definitive proof that the chosen features reflect the classifier's inner workings entirely accurately, we decided that using those mechanisms for the feature choice are the best proxy that we have.
As such, after the participants went through the eight examples that were used for the prediction task, they were asked for each feature how much they agreed that it was relevant to the AI's decisions on a 5-point Likert scale (0~=~strongly disagree, 4~=~strongly agree).
%They also had to provide a short free text justification about why they think that the AI decides after these criteria.
%To aid them in their task, they were again shown the eight example images from the previous prediction task together with the classifier's decisions and the explanations corresponding to their condition.

\paragraph{Explanation Satisfaction}
In order to measure the participants' subjective satisfaction, we used the Explanation Satisfaction Scale proposed by \citet{hoffman2018metrics} which consists of eight items rated on a 5-point Likert scale (0~=~strongly disagree, 5~=~strongly agree) that we averaged over all items. Since it does not apply to our use-case, we excluded the 5th question of the questionnaire. The seven remaining items address \emph{confidence}, \emph{predictability}, \emph{reliability}, \emph{safety}, \emph{wariness}, \emph{performance}, \emph{likeability}.
Finally, the participants had the possibility to give free text feedback.

\subsection{Participants}
%\begin{figure}
%    \centering
%    % \captionsetup{justification=raggedright,margin=.5cm}
%    \includegraphics[width=.4\linewidth]{figures/percentage_correct_ai_predictions.png}
%     \caption{Mean participant prediction accuracy of the AI's prediction by condition. The conditions containing alterfactual explanations outperformed all other conditions. Error bars represent the 95\% CI. *\textit{p}~$<$~.05,  **\textit{p}~$<$~.001.}
%    \label{fig:prediction_accuracy}
%\end{figure}

\begin{figure}
    \centering
    % \captionsetup{justification=raggedright,margin=.5cm}
    \begin{minipage}{.42 \linewidth}
        \includegraphics[width=\linewidth]{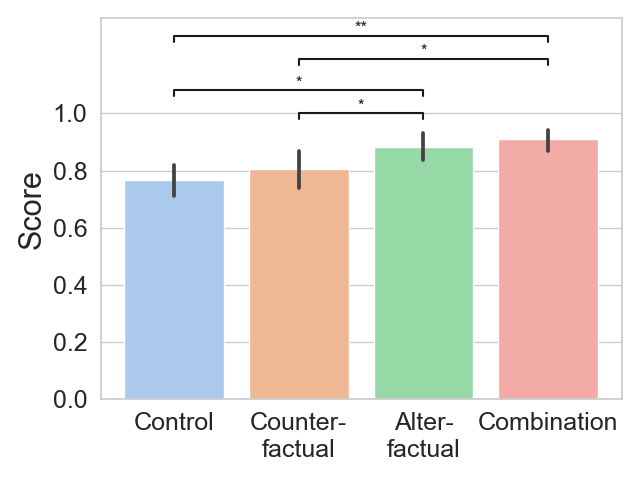}
        \centering
    \end{minipage}
    \begin{minipage}{.42 \linewidth}
        \includegraphics[width=\linewidth]{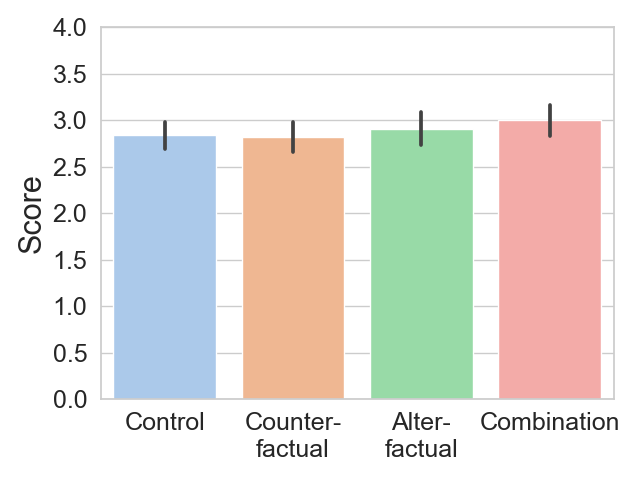}
        \centering
    \end{minipage}
     \caption{Left: Mean participant prediction accuracy of the AI's prediction by condition. The conditions containing alterfactual explanations outperformed all other conditions. 
     Right: Mean understanding of the irrelevant and relevant features in our study. 
     Error bars represent the 95\% CI. *\textit{p}~$<$~.05,  **\textit{p}~$<$~.001.}
    \label{fig:prediction_accuracy}
\end{figure}
%Insert information about Power Analysis. 
% Trough a power analysis, we estimated a required sample size of n~=~30 per condition for a MANOVA with 80\% power and an alpha of 5\%. Therefore, we assumed a medium effect as it was reported for the study by Mertes et al. \cite{mertes2022alterfactual}.
Through a power analysis, we estimated a required sample size of at least 21 per condition for a MANOVA with 80\% power and an alpha of 5\%, based on the Pillai’s Trace of 0.13 reported for the study by \citet{mertes2022alterfactual}.
131 Participants between 18 and 29 years (\textit{M}~=~22.2, \textit{SD}~=~2.44) were recruited at the University of \emph{blinded for review}. 
61 of them were male, 70 female. 
The participants were randomly separated into the four conditions (33 per condition and 32 in the Alterfactual condition). 
%Subjects of the three explanation conditions that did not look at a single explanation during the whole study were excluded from the evaluation. There were no such cases. 
The highest level of education that most participants held (76.3\%) was a high-school diploma.
Only 11.5\% of the participants had no experience with AI. Most of the participants (74\%) have heard from AI in the media. 
Excluding participants that had no opinion on the subject, the participants expected a positive impact of AI systems in the future (\textit{M}~=~3.73 on a 5-point Likert Scale from 1~=~"Extremely negative" to 5~=~"Extremely positive").
There were no substantial differences in the demographics between conditions (see appendix). 
%Participants in the Alterfactual condition had a lower rate of females and participants in the Control condition had a less positive attitude towards AI. 

\section{Results}

%% TODO this section is very similar to the last alterfactual paper, maybe check for plagiarism

\subsection{Model Understanding}
%To investigate the impact of the four different experimental conditions\footnote{(\Control{} condition, \Counterfactual{} condition, \Alterfactual{} condition, \Combination{} condition)} on the (1) feature understanding and (2) prediction accuracy, we conducted a MANOVA. 
To investigate the impact of the four different experimental conditions on the (1) feature understanding and (2) prediction accuracy, we conducted a MANOVA. 
We found a significant difference, Wilks' Lambda~=~0.859, \textit{F}(6,252)~=~3.31, \textit{p}~=~.004. 

The following ANOVA revealed that \textbf{only the prediction accuracy of the participants showed significant differences between the conditions}:
\begin{itemize}
    \item \emph{Feature Understanding}: \textit{F}(3,127)~=~0.877, \textit{p}~=~.455.
    \item \emph{Prediction Accuracy}: \textit{F}(3,127)~=~6.578, \textit{p}~$<$~.001.
\end{itemize}

%\begin{figure}
%    \centering
%    % \captionsetup{justification=raggedright,margin=.5cm}
%    \begin{minipage}{.43 \linewidth}
%        \includegraphics[width=\linewidth]{figures/irrelevant_understanding.png}
%        \centering
%        Irrelevant Features
%    \end{minipage}
%    \begin{minipage}{.43 \linewidth}
%        \includegraphics[width=\linewidth]{figures/relevant_understanding.png}
%        \centering
%        Relevant Features
%    \end{minipage}
%     \caption{Mean understanding of the irrelevant and relevant features in our study. Error bars represent the 95\% CI.}
%    \label{fig:relevant_and_irrelevant}
%\end{figure}
As displayed in Figure \ref{fig:prediction_accuracy}, the post-hoc t-tests showed that the participants' prediction accuracy was significantly better in the \Alterfactual{} and \Combination{} conditions compared to the other conditions. The effect size \textit{d} is calculated according to \citet{cohen2013statistical}:%\footnote{Interpretation of the effect size is: \textit{d}~$<$~.5~:~small effect; \textit{d}~=~0.5-0.8~:~medium~effect; \textit{d}~$>$~0.8~:~large~effect}:

\begin{itemize}
    \item \textbf{\Alterfactual{} vs. \Control{}}: \textit{t}(127)~=~3.19, \textit{p}~=~.002, \textit{d}~=~0.79 (medium effect).
    \item \textbf{\Alterfactual{} vs. \Counterfactual{}}: \textit{t}(127)~=~2.06, \textit{p}~=~.042, \textit{d}~=~0.51 (medium effect).
    \item \textbf{\Combination{} vs. \Control{}}: \textit{t}(127)~=~3.93, \textit{p}~$<$~.001, \textit{d}~=~0.97 (large effect).
	\item \textbf{\Combination{} vs. \Counterfactual{}}: \textit{t}(127)~=~2.79, \textit{p}~=~.006, \textit{d}~=~0.69 (medium effect).
\end{itemize}

These results regarding the prediction task confirm our hypothesis that the \textbf{conditions with alterfactual explanations outperform the condition without explanations in the prediction task.}
Further, \textbf{the combination of both explanation types did significantly outperform counterfactual explanations.}
However, our hypothesis that the combination is more effective in terms of enabling a correct model understanding than alterfactual explanations has to be rejected. 
%as well as our hypotheses that alterfactual and counterfactual explanations help with the feature relevance task have to be rejected.

\subsection{Relevant and Irrelevant Information}
As reported in the section above, we did not find a significant overall difference in the feature understanding task (see Figure \ref{fig:prediction_accuracy}). 
However, in order to investigate our hypotheses about irrelevant vs. relevant features, we conducted another MANOVA between the conditions and the combined understanding values for the two relevant features and the two irrelevant features.
This MANOVA did not find any significant differences, Wilks' Lambda~=~0.951, \textit{F}(6,252)~=~1.07, \textit{p}~=~.379.
The mean understanding per condition can be found in the appendix. %in Figure~\ref{fig:relevant_and_irrelevant}.

\subsection{Explanation Satisfaction}
The ANOVA revealed that there were no significant differences in the subjective explanation satisfaction between the three explanation conditions, \textit{F}(2,95)~=~0.34, \textit{p}~=~.713.
The mean satisfaction values with standard deviation were: \Counterfactual{} condition: $3.54\pm 0.53$ ; \Alterfactual{} condition: $3.65 \pm 0.6$; \Combination{} condition: $3.58\pm 0.5$.

% \begin{figure}
%     \centering
%      \captionsetup{justification=raggedright,margin=.5cm}
%     \includegraphics[width=.5\linewidth]{figures/average_satisfaction.png}
%      \caption{Mean explanation satisfaction by condition. Error bars represent the 95\% CI.}
%     \label{fig:explanation_satisfaction}
%\end{figure}

\section{Discussion}
%In the user study, we investigated the impact of our four different experimental conditions, namely \Control{}, \Counterfactual{}, \Alterfactual{}, and \Combination{}, on two key aspects: (1) model understanding and (2) explanation satisfaction. Thereby, we assessed the users' local understanding of the classifier by a proxy prediction task, whereas the global understanding was evaluated by directly asking about the feature importance of certain relevant or irrelevant features.
With our proposed GAN-based approach, we demonstrated that it is possible to generate both counterfactual and alterfactual explanations for a black box image classifier. Using computational metrics, we showed that both of those generated explanations fulfill their respective requirements: The counterfactual explanations are very similar to the original images (i.e., 0.90 average SSIM) but change the classifiers prediction in 87.70\% of the cases while alterfactual explanations are very different from the original image (i.e., 0.32 average SSIM), but do not change the classifier's prediction in 96.20\% of the cases.
%While for the prediction task of our user study, alterfactual explanations and the combination of alterfactual and counterfactual explanations performed significantly better than the other two conditions, we did not observe a significant difference for the feature relevance understanding.
For the prediction task of our user study, alterfactual explanations and the combination of alterfactual and counterfactual explanations performed significantly better than the other two conditions demonstrating the potential of alterfactual explanations to facilitate local model understanding.
However, we did not observe a significant difference for the feature relevance understanding.
This is highly interesting, as it contrasts with a previous study by \citet{mertes2022alterfactual}. There, a similar experimental design was employed for assessing the effect that alterfactual explanations have on users' mental models of a hard-coded classifier that assesses numerical feature descriptors for a fictional classification problem.
In contrast to our work, they neither used a real classifier nor an alterfactual generation algorithm, but only mock-up decisions and explanations.
In their scenario, alterfactual explanations led to a significantly better feature relevance understanding, while not having a substantial impact on the performance in a prediction task.
A possible explanation for this is the fact that our study was conducted in the context of fashion classification, where
%, where we used an image classifier in the context of fashion classification, 
the users might already have had a quite distinctive mental model of the problem domain itself. 
Further, images might be more accessible than numerical feature descriptors to end users.
As such, the global understanding of the classifier might already be positively biased. This argument is supported by looking at the feature relevance understanding results of the control group - although not seeing any explanations, they already performed very well in identifying relevant features.
%However, as can be seen by the significant performance improvement in the prediction task, the local understanding of the model does not benefit from that, as the classification model is not perfect - a global understanding of the use case itself does not necessarily imply an understanding of cases where the classifier's decision does not correctly model the reality. 
However, as can be seen by the significant performance improvement in the prediction task, the local understanding of the model does not necessarily benefit from the identification of globally relevant features.
%benefit from this effect.
As the classification model is imperfect, a global understanding of the use case itself does not necessarily imply an understanding of cases, e.g. when the classifier's decision does not correctly model reality. 
Furthermore, our results regarding the global model understanding should be taken with a grain of salt, since the fashion-classifier is a black-box model - even though we used post-hoc explanation methods (SHAP and LIME), we cannot be certain that our choice of features is completely accurate.
%to find a basis for the feature relevance understanding. Such post-hoc methods can be misleading in some situtaions \cite{rudin, tobi}.
%to evaluate if the answers of the participants were correct - those mechanisms have their own flaws, which is why we cannot be sure that our feature choice is completely accurate.
%The effect sizes for the prediction task obtained from the post-hoc t-tests indicate medium to large effects, which strengthens the practical significance of the observed differences. 
%The effect sizes suggest that the improvements in prediction accuracy are not only statistically significant but also meaningful in real-world scenarios.
Interestingly, we did not observe any significant differences in explanation satisfaction. This indicates that participants felt similarly satisfied by all explanation methods even though the alterfactual and combined explanations objectively helped more during the prediction task. 
The presentation of more information (i.e., in the combination condition) could have led to a higher cognitive load and influenced the subjective assessments of explanation satisfaction, resulting in the difference between objective measurement (i.e., model understanding) and subjective measurement (i.e., explanation satisfaction).

\section{Conclusion}
% In this paper, we studied a recently proposed concept for explaining AI models called \emph{alterfactual explanations} that \emph{alter} as much irrelevant information as possible while maintaining the distance to the decision boundary.
% We demonstrated for the first time that it is in fact feasible to generate such explanations for black box models and briefly evaluated them computationally.
In this paper, we demonstrate the practical feasibility of a recently proposed concept for explaining AI models called \emph{alterfactual explanations} that \emph{alter} as much irrelevant information as possible while maintaining the distance to the decision boundary.

We show for the first time that it is possible to generate such explanations for black box models and briefly evaluated them computationally.
Furthermore, we showed in a user study that our generated alterfactual explanations can complement counterfactual explanations.
%In this paper, we introduced the novel explanation method of \emph{Alterfactual Explanations}.
%A GAN-based approach for generating such explanations was presented and evaluated in a user study.
In that study, we compared how users' model understanding of a binary image classifier changes when being confronted with counterfactual explanations, alterfactual explanations, or a combination of both. Further, a control group was assessed that did not see any explanations.
We found that in a prediction task, where the classifier's prediction had to be anticipated by looking at the explanations, users performed significantly better when they were provided with explanations that included alterfactual explanations compared to users that did not see alterfactual explanations, although we did not observe a significant difference in explanation satisfaction.
%Our user study also identified that alterfactual explanations did not increase participants' explanation satisfaction, although they significantly improved the users' performance in the prediction task.
%Our user study also identified that, although significantly improving the users' performance in the prediction task, alterfactual explanations did not increase participants' explanation satisfaction.
%As such, future work should investigate why this is the case, e.g., by studying the approach in other use cases and scenarios.
%All in all, we can confidently claim that the concept of alterfactual explanations is a promising approach to make AI more transparent and understandable to end users.

Overall, we showed that alterfactual explanations are a promising explanation method that can complement counterfactual explanations in future XAI systems.
\clearpage
\section{Acknowledgments}
This research was partially funded by the DFG through the Leibniz award of Elisabeth André (AN 559/10-1).

%% The file named.bst is a bibliography style file for BibTeX 0.99c
\bibliographystyle{named}
\bibliography{main}

\appendix
\section{GAN Architecture and Training}
\subsubsection{Generator Model}
The GAN's generator architecture is listed in Table \ref{table:gen}.

\subsubsection{Discriminator Model}
The GAN's discriminator architecture is listed in Table \ref{table:d1}.

\subsection{Training Configuration and Hyperparameters}
The training configuration and hyperparamters are shown in Table \ref{table:gan_configs}. The Adam optimizer was configured with $\beta_{1} = 0.5$, $\beta_{2} = 0.999$, $\epsilon = 1e-8$.

Further, the Support Vector Machine (SVM) that was included in the loss function (see main paper) was trained with the parameters listed in Table \ref{table:svm_configs}.

\section{Classifier Architecture and Training}
In Table \ref{table:cl1}, the model architecture for the classifier that we used in our evaluation scenario is described. 
The training configuration and hyperparamters are shown in Table \ref{table:classifier_configs}. The Adam optimizer was configured with $\beta_{1} = 0.9$, $\beta_{2} = 0.999$, $\epsilon = 1e-8$.

\section{Additional Dataset Experiments}
In order to demonstrate that our alterfactual generation approach is generalizable to different datasets, we additionally trained models for three other datasets.
Here, we omitted the Feature Relevance component.
As for that component an additional SVM has to be trained on the penultimate layer of the classifier layer, it takes away the model-agnostic property from the alterfactual generation network.
By performing these additional experiments, we show that the approach can simply be adapted to be model-agnostic, although that may negatively affect the outcomes of the results - it is not specifically forced that \emph{only} irrelevant features change.
For the classifiers, we used the same architecture as for the Fashion-MNIST dataset, although batch size and epochs were modified to fit the hardware that we used.

\subsection{MNIST}
\begin{figure*}
    \centering
    \includegraphics[width=\textwidth]{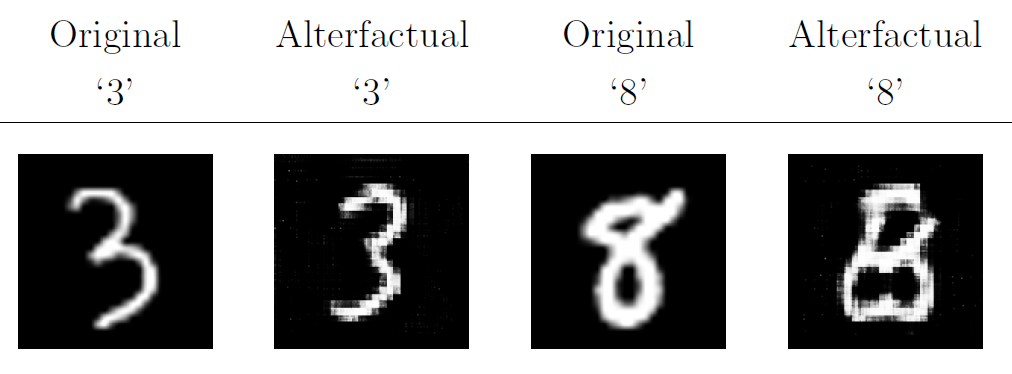}
    \caption{Examplary alterfactual outputs for the MNIST dataset.}
    \label{fig:mnist_example}
\end{figure*}

As the MNIST datasets has more than two classes  (each class contains hand-drawn images of one specific digit), we picked the two digits that are most likely to be confused by deep learning classifiers: \emph{Three} and \emph{Eight}.
The MNIST classifier was trained for 9 epochs with batch size 32.
Besides not using the Feature Relevance component and increasing the epoch number to 42, the GAN network was trained with the same parameter settings as for the Fashion-MNIST dataset.
We reached a validity of 95.92\% and an average SSIM of 0.425.
Example outputs are shown in Figure \ref{fig:mnist_example}.

{}

\subsection{MaskedFace-Net}
\begin{figure*}
    \centering
    \includegraphics[width=\textwidth]{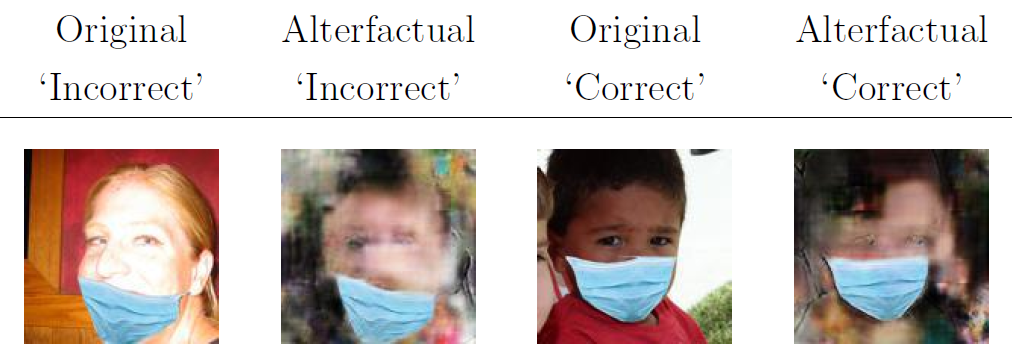}
    \caption{Examplary alterfactual outputs for the MaskedFace-Net dataset.}
    \label{fig:mask_example}
\end{figure*}

The MaskedFace-Net dataset contains images of people wearing face masks. Binary labels are provided, indicating that on the respective image the mask is worn correctly or incorrectly.
The classifier was trained for 2 epochs with batch size 128.
Besides not using the Feature Relevance component and decreasing the epoch number to 11, the GAN network was trained with the same parameter settings as for the Fashion-MNIST dataset.
We reached a validity of 84.27\% and an average SSIM of 0.091.
Example outputs are shown in Figure \ref{fig:mask_example}.

{}

\subsection{MaskedFace-Net (Gray Scale)}
\begin{figure*}
    \centering
    \includegraphics[width=\textwidth]{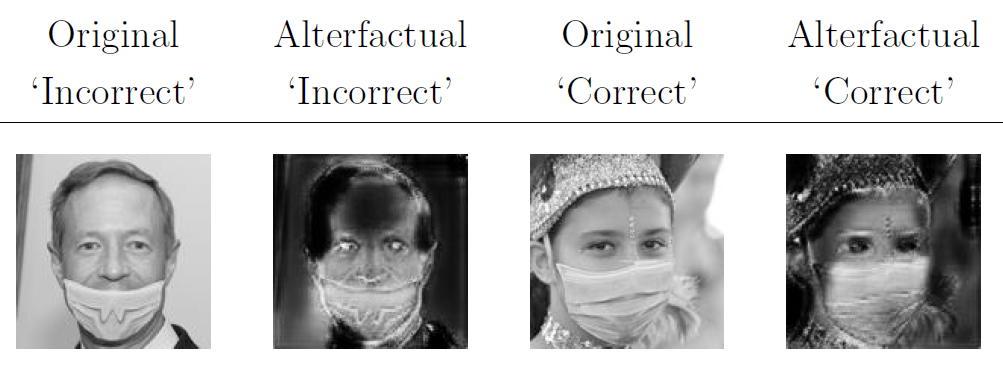}
    \caption{Examplary alterfactual outputs for the gray scale version of the MaskedFace-Net dataset. It can be clearly seen that the only part of the image that gets unchanged is the mask itself - indicating that everything else is irrelevant.}
    \label{fig:maskg_example}
\end{figure*}

Here, we also used the MaskedFace-Net dataset, but converted it to gray scale, demonstrating that our approach also works with gray scale data. 
The classifier was trained for 1 epoch with batch size 128.
Besides not using the Feature Relevance component and decreasing the epoch number to 6, the GAN network was trained with the same parameter settings as for the Fashion-MNIST dataset.
We reached a validity of 48.89\% and an average SSIM of 0.002.
Example outputs are shown in Figure \ref{fig:maskg_example}.

{}

\section{User Study}
\label{ap:demographics}

\subsection{Demographic Details}

\begin{figure}[h]
    \centering
    \small
    \begin{minipage}{0.48\linewidth}
    \centering
    \includegraphics[width=\linewidth]{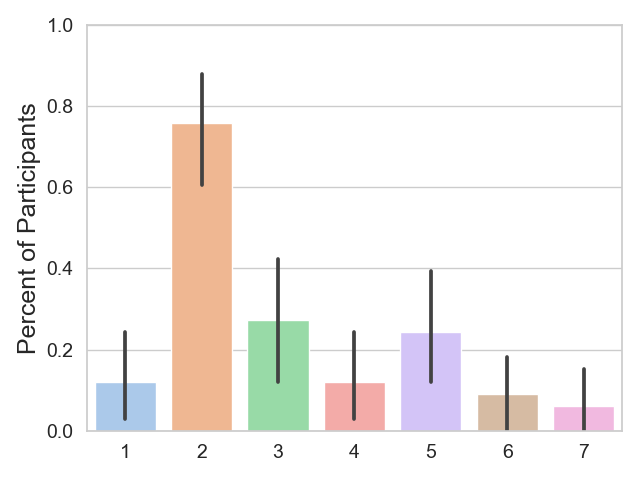}
    \Control{}
    \end{minipage}
        \begin{minipage}{0.48\linewidth}
    \centering
    \includegraphics[width=\linewidth]{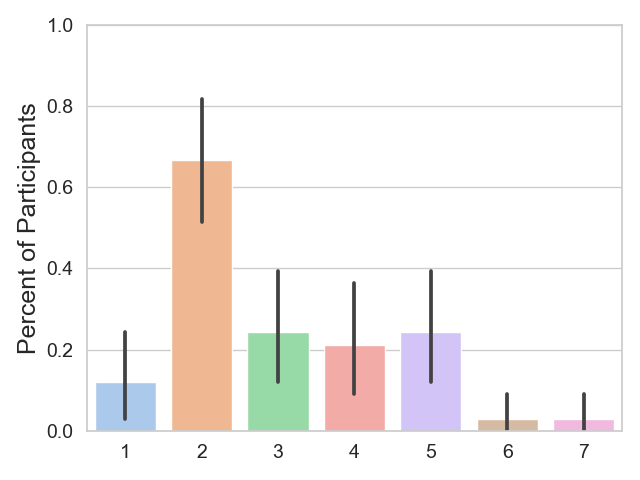}
    \Counterfactual{}
    \end{minipage}
    
    \begin{minipage}{0.48\linewidth}
    \centering
    \includegraphics[width=\linewidth]{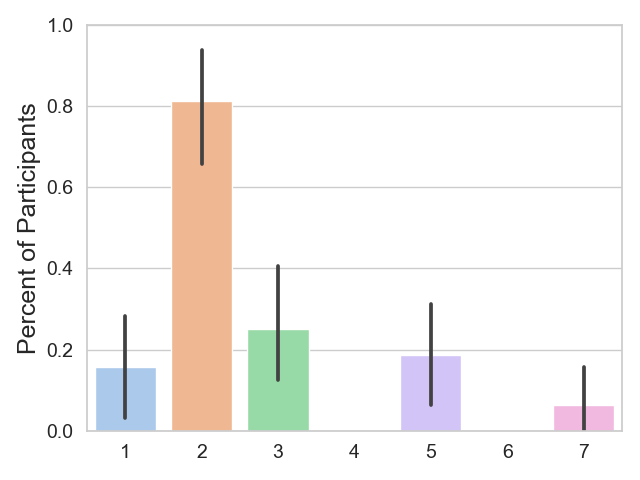}
    \Alterfactual{}
    \end{minipage}
        \begin{minipage}{0.48\linewidth}
    \centering
    \includegraphics[width=\linewidth]{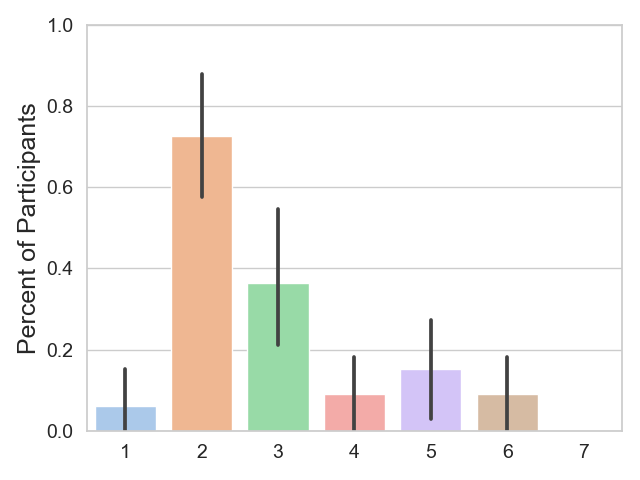}
    \Combination{}
    \end{minipage}
    \caption{Distribution of the chosen AI experience items for each condition. The x-axis depicts the following items: 1 - I do not have any experience in AI related topics; 2 - I know AI from the media; 3 - I use AI technology in my private life; 4 - I use AI technology in my work; 5 - I have taken at least one AI related course; 6 - I do research on AI-related topics; 7 - Other: }
    \label{fig:experience_XAI}
\end{figure}{}
The mean age and education level, as well as the percentage of female participants, per condition can be seen in Table \ref{tab:demographics_per_condition}.
For the AI experience and Attitude we adapted a description of AI from \citet{zhang2019artificial} and \citet{russell2016artificial} to ``The following questions ask about Artificial Intelligence (AI). Colloquially, the term `artificial intelligence' is often used to describe machines (or computers) that mimic `cognitive' functions that humans associate with the human mind, such as `learning' and `problem solving'.'' 
After this description, participants had to select one or more item describing their experience with AI.
The distribution of the items for each condition is shown in Fig. \ref{fig:experience_XAI}.
Following this we adapted a question from \citet{zhang2019artificial} to measure the participants' attitude towards AI. We asked them to rate their answer to the question ``Suppose that AI agents would achieve high-level performance in more areas one day. How positive or negative do you expect the overall impact of such AI agents to be on humanity in the long run?'' on a 5-point Likert scale from ``Extremely negative'' to ``Extremely positive''. The participants also had the option to answer ``I do not know'' here, which would exclude them from the evaluation of this question. The mean results for each condition are shown in Table \ref{tab:demographics_per_condition}.

\section{Additional Post-Hoc Results}

For completeness, we also report the results of the post-hoc t-tests on the participants' prediction accuracy that were not significant. The effect size \textit{d} is calculated according to \citet{cohen2013statistical}:%\footnote{Interpretation of the effect size is: \textit{d}~$<$~.5~:~small effect; \textit{d}~=~0.5-0.8~:~medium~effect; \textit{d}~$>$~0.8~:~large~effect}:

\begin{itemize}
    \item \textbf{\Counterfactual{} vs. \Control{}}: \textit{t}(127)~=~1.14, \textit{p}~=~.258, \textit{d}~=~0.28
    \item \textbf{\Combination{} vs. \Alterfactual{}{}}: \textit{t}(127)~=~0.71, \textit{p}~$<$~.478, \textit{d}~=~0.18.
\end{itemize}

For feature understanding and explanation satisfaction we did not calculate post-hoc tests since the ANOVA was not significant.

\subsection{Mean Understanding of Features}
\begin{figure}
    \centering
    % \captionsetup{justification=raggedright,margin=.5cm}
    \begin{minipage}{.43 \linewidth}
        \includegraphics[width=\linewidth]{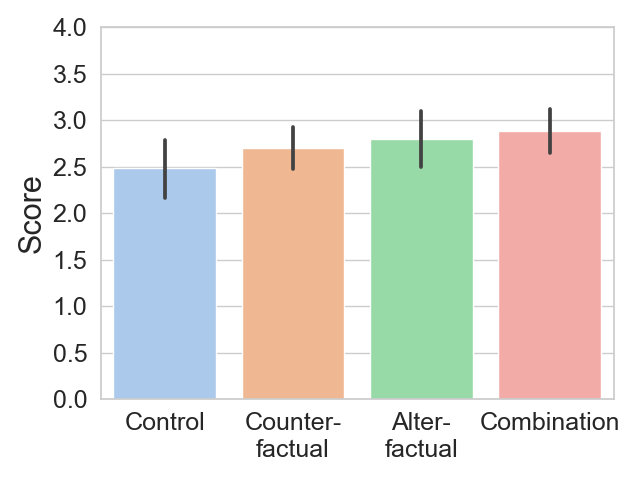}
        \centering
        Irrelevant Features
    \end{minipage}
    \begin{minipage}{.43 \linewidth}
        \includegraphics[width=\linewidth]{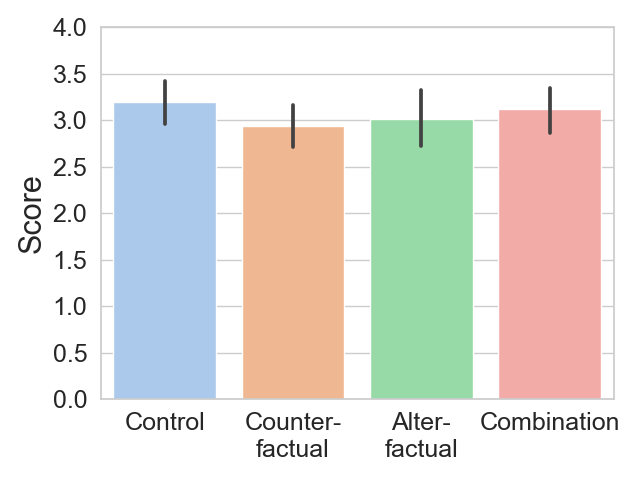}
        \centering
        Relevant Features
    \end{minipage}
     \caption{Mean understanding of the irrelevant and relevant features in our study. Error bars represent the 95\% CI.}
    \label{fig:relevant_and_irrelevant}
\end{figure}

Figure \ref{fig:relevant_and_irrelevant} shows the mean understanding (as assessed by the feature understanding task) per condition.

\subsection{Explanation Satisfaction Scale}
For evaluating explanation satisfaction, we used the Explanation Satisfaction scale by Hoffmann \citep{hoffman2018metrics} except one item that did not apply to our use case. The items that we used were as follows, where each item was rated on a 5-point likert scale (1 = strongly disagree, 5 = strongly agree):
\begin{itemize}
    \item From the explanations, I \textbf{understand} how the AI makes its decision.
    \item The explanations of how the AI makes its decision are \textbf{satisfying}.
    \item The explanations of how the AI makes its decision have \textbf{sufficient detail}.
    \item The explanations of how the AI makes its decision seem \textbf{complete}.
    \item The explanations of how the AI makes its decision are \textbf{useful} to predict the AI's decision.
    \item The explanations of how the AI makes its decision show me how \textbf{accurate} the AI is.
    \item The explanations let me judge when I should \textbf{trust and not trust} the AI.
\end{itemize}

\subsection{Study Design}
\begin{figure*}
    \centering
    \includegraphics[width=\paperwidth]{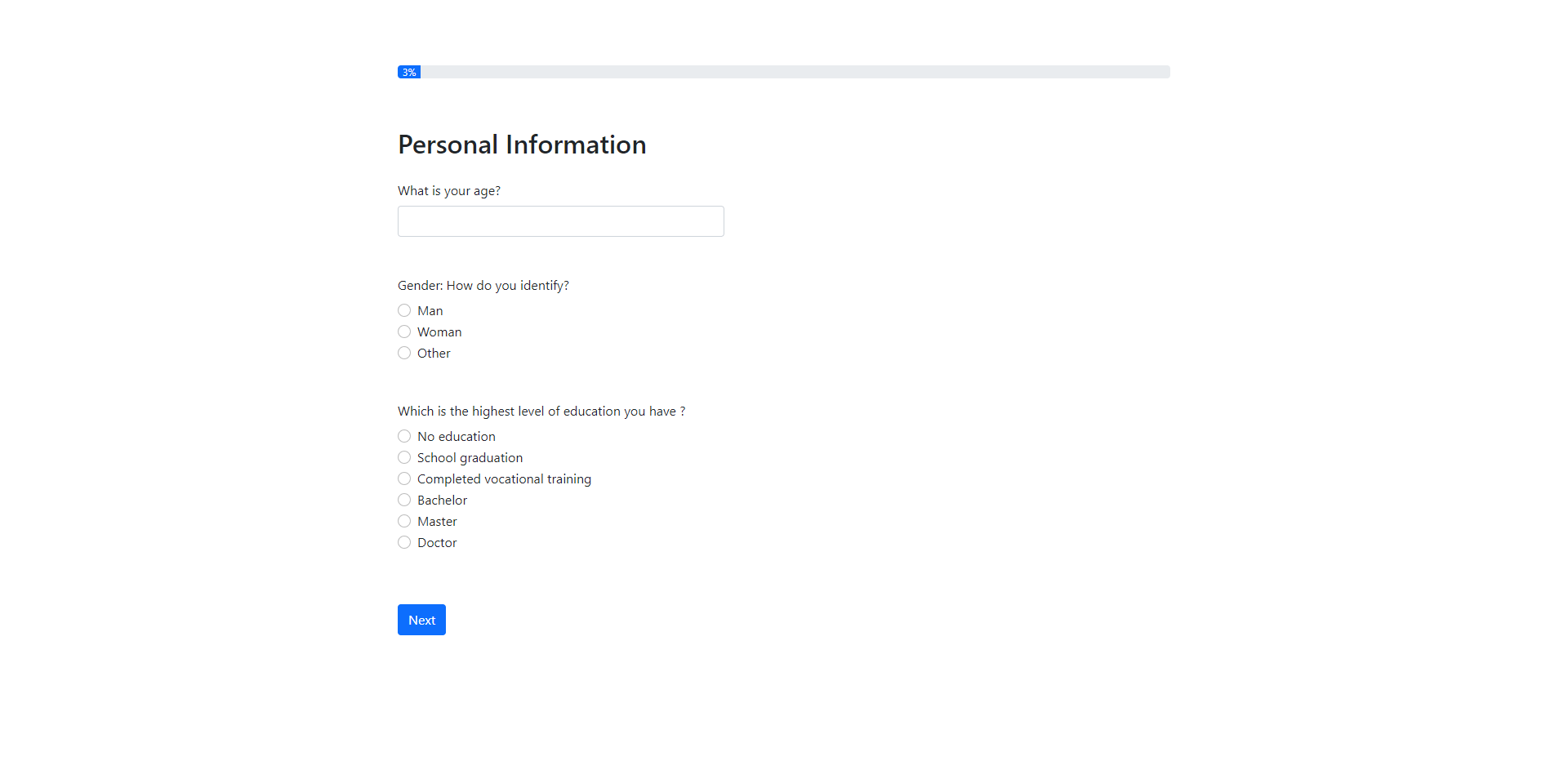}
    \caption{Screenshot of the user study, part 1.}
    \label{fig:study_screenshot_1}
\end{figure*}

\begin{figure*}
    \centering
    \includegraphics[width=\paperwidth]{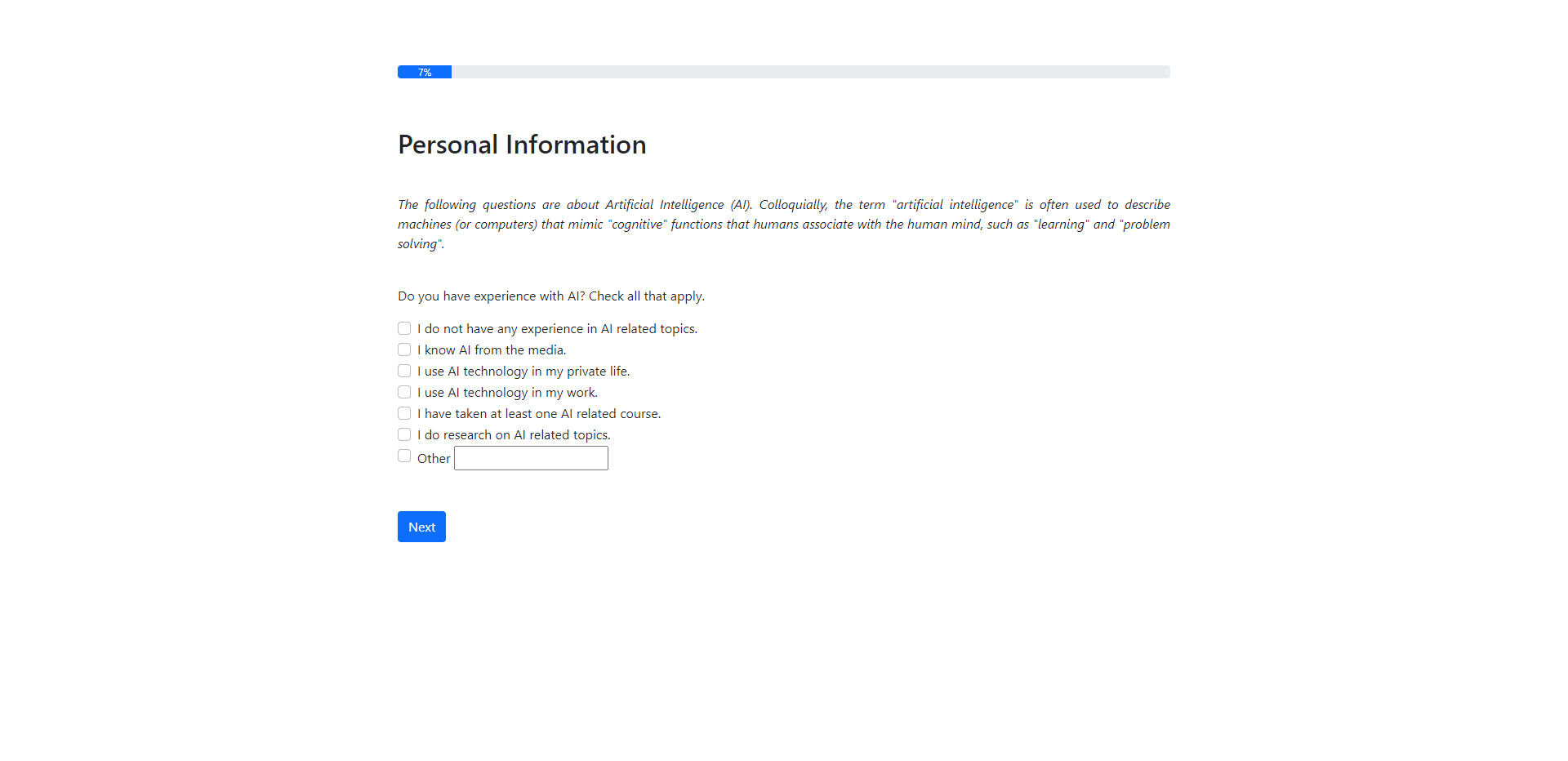}
    \caption{Screenshot of the user study, part 2.}
    \label{fig:study_screenshot_2}
\end{figure*}

\begin{figure*}
    \centering
    \includegraphics[width=\paperwidth]{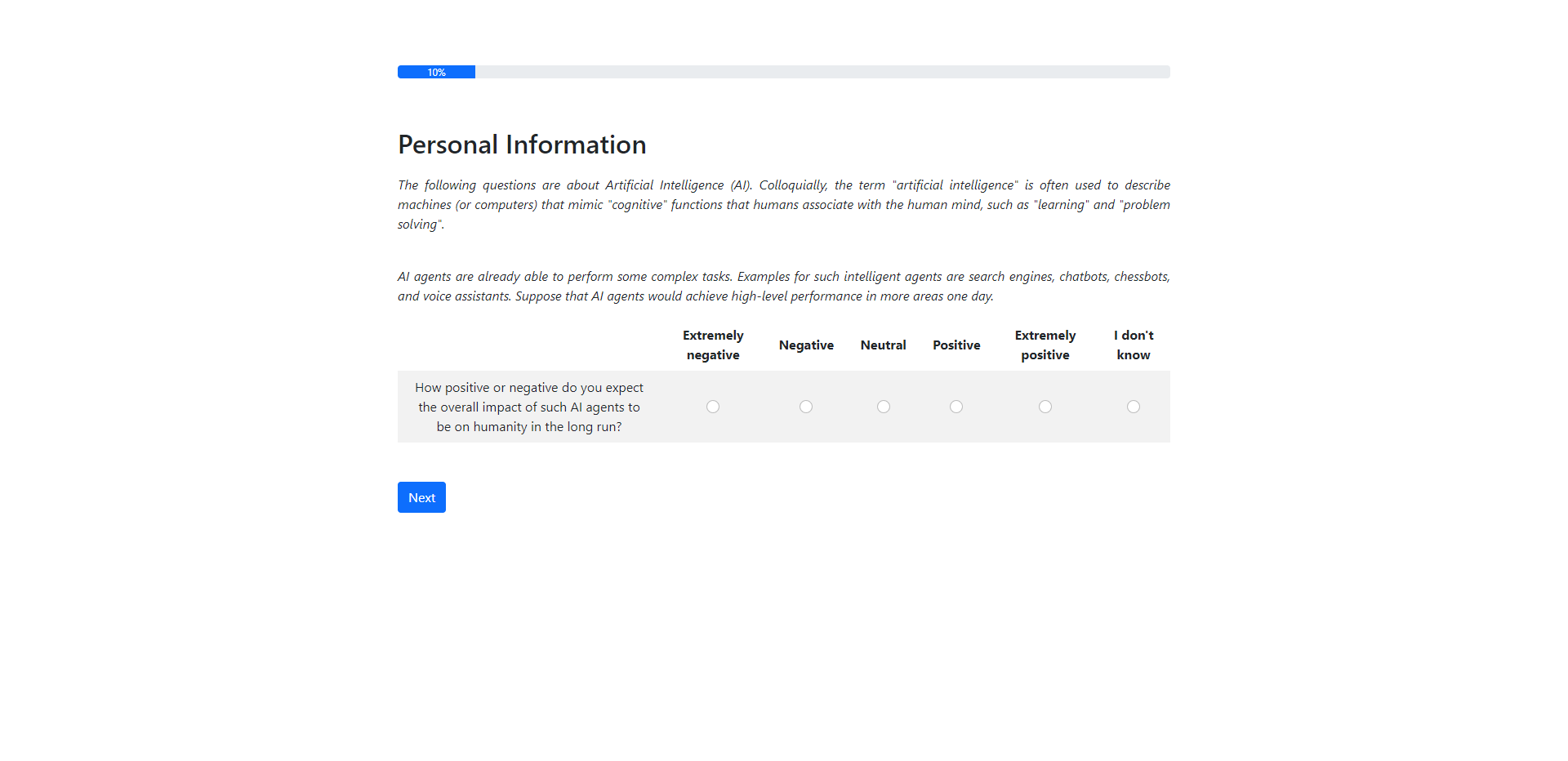}
    \caption{Screenshot of the user study, part 3.}
    \label{fig:study_screenshot_3}
\end{figure*}

\begin{figure*}
    \centering
    \includegraphics[width=\paperwidth]{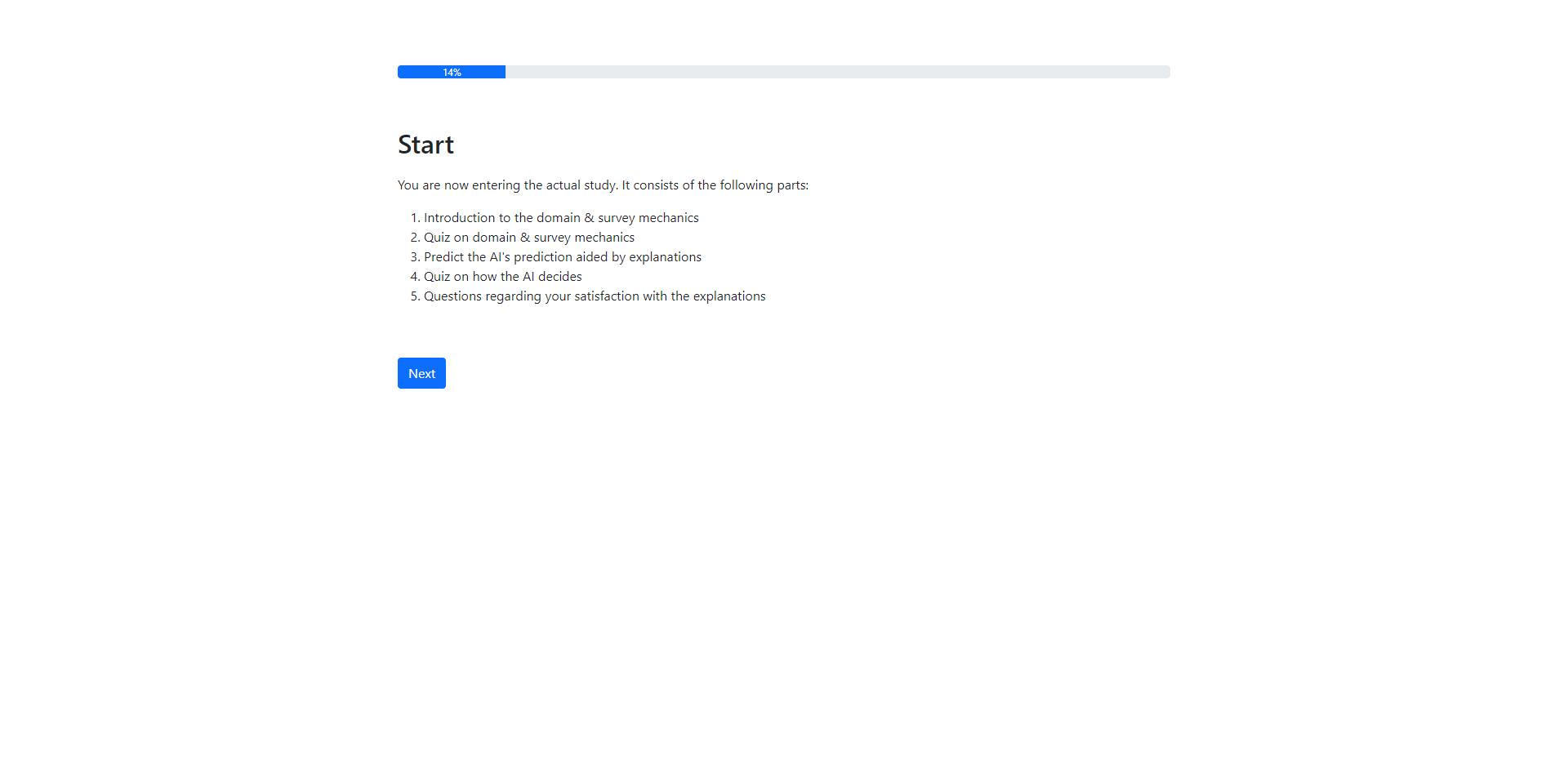}
    \caption{Screenshot of the user study, part 4.}
    \label{fig:study_screenshot_4}
\end{figure*}

\begin{figure*}
    \centering
    \includegraphics[width=\paperwidth]{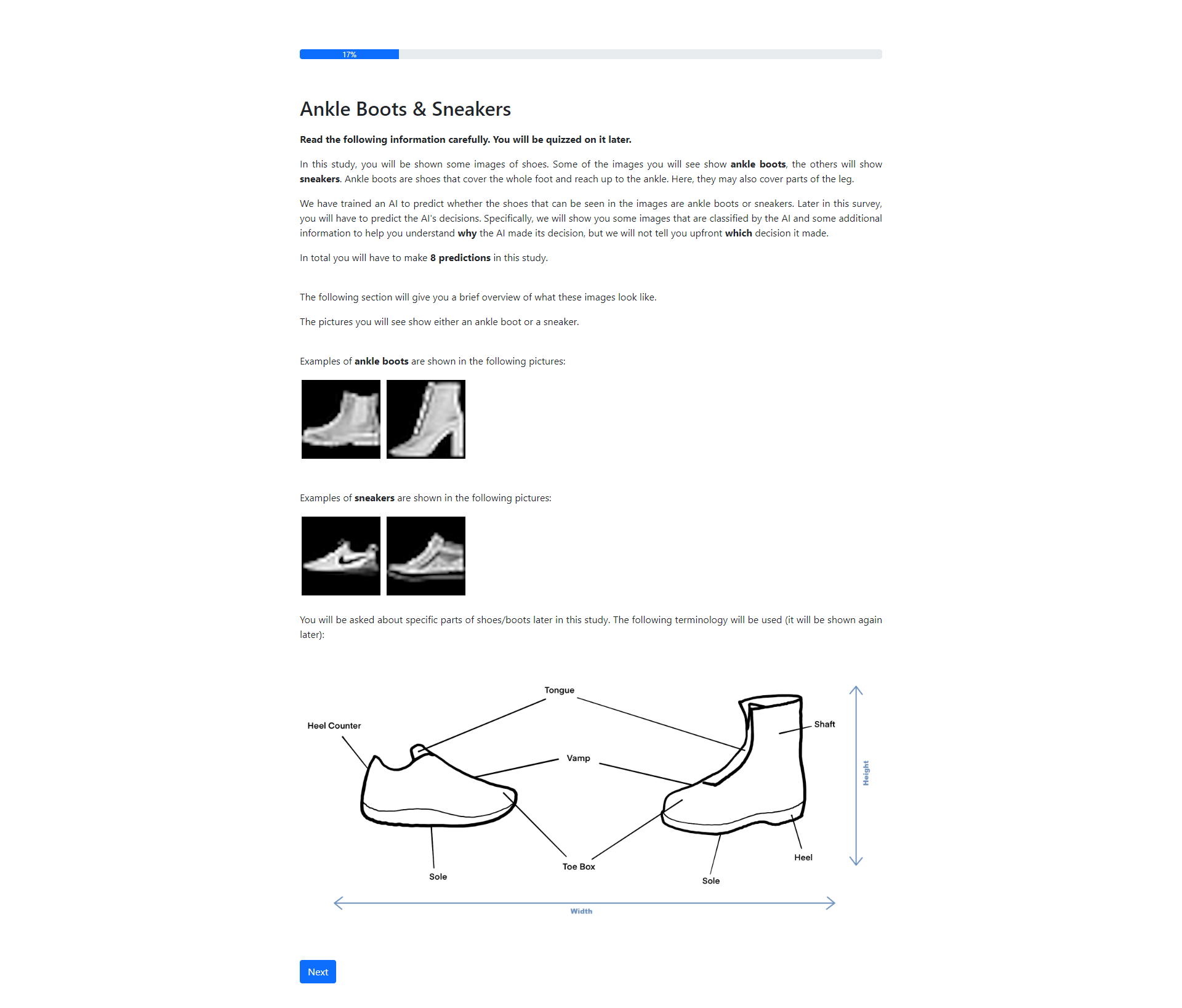}
    \caption{Screenshot of the user study, part 5.}
    \label{fig:study_screenshot_5}
\end{figure*}

\begin{figure*}
    \centering
    \includegraphics[width=\paperwidth]{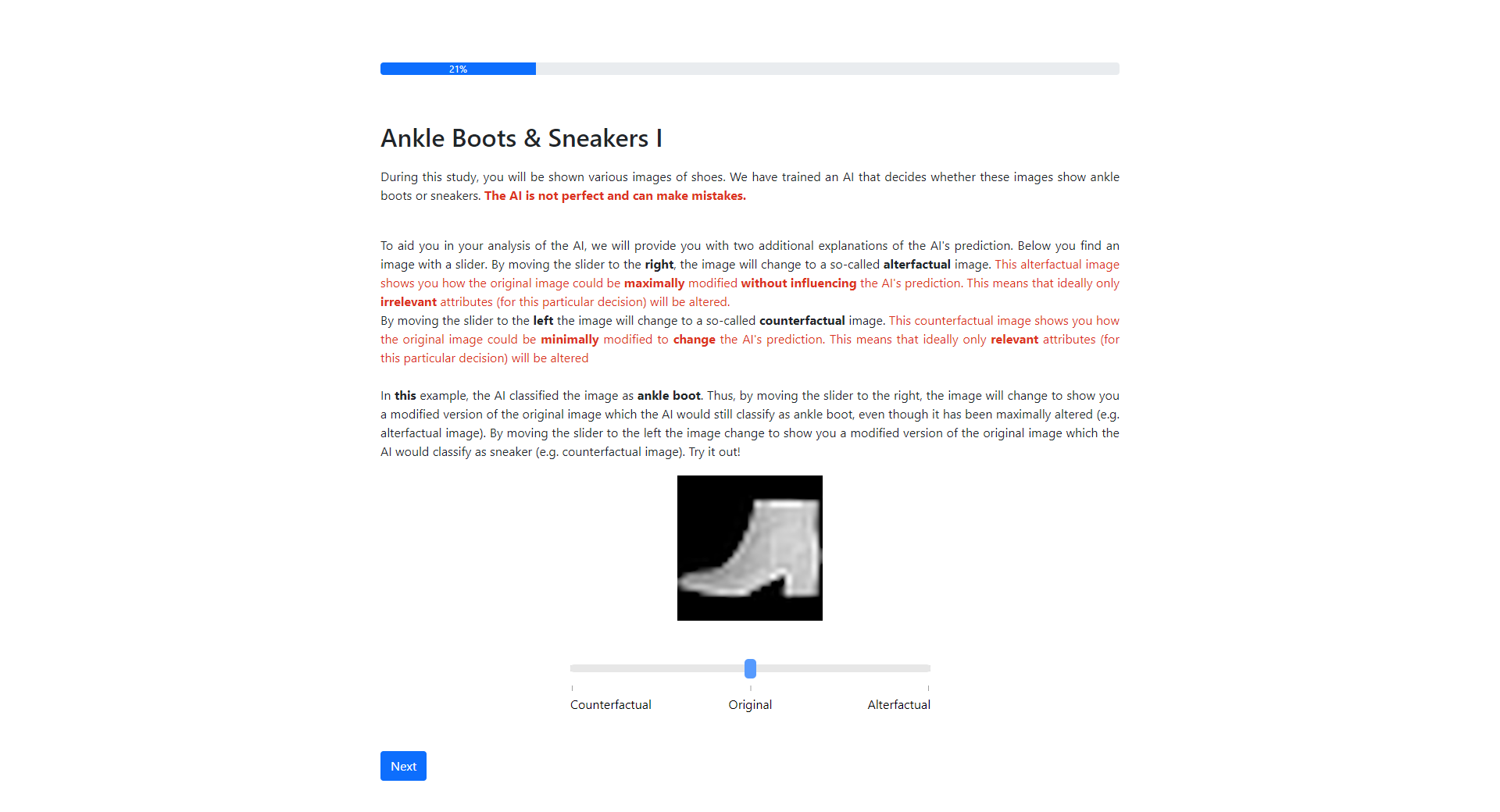}
    \caption{Screenshot of the user study, part 6.}
    \label{fig:study_screenshot_6}
\end{figure*}

\begin{figure*}
    \centering
    \includegraphics[width=\paperwidth]{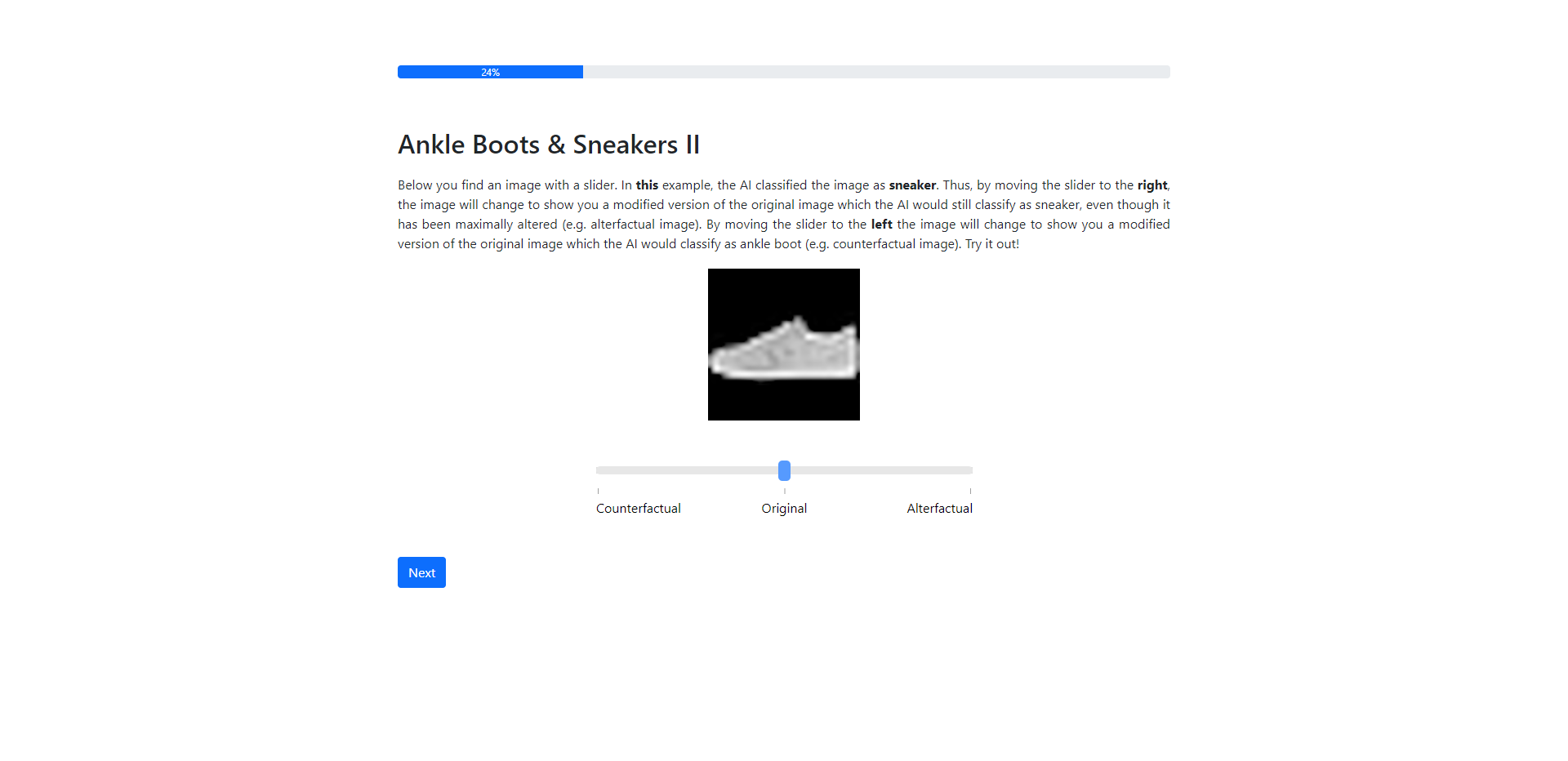}
    \caption{Screenshot of the user study, part 7.}
    \label{fig:study_screenshot_7}
\end{figure*}

\begin{figure*}
    \centering
    \includegraphics[width=\paperwidth]{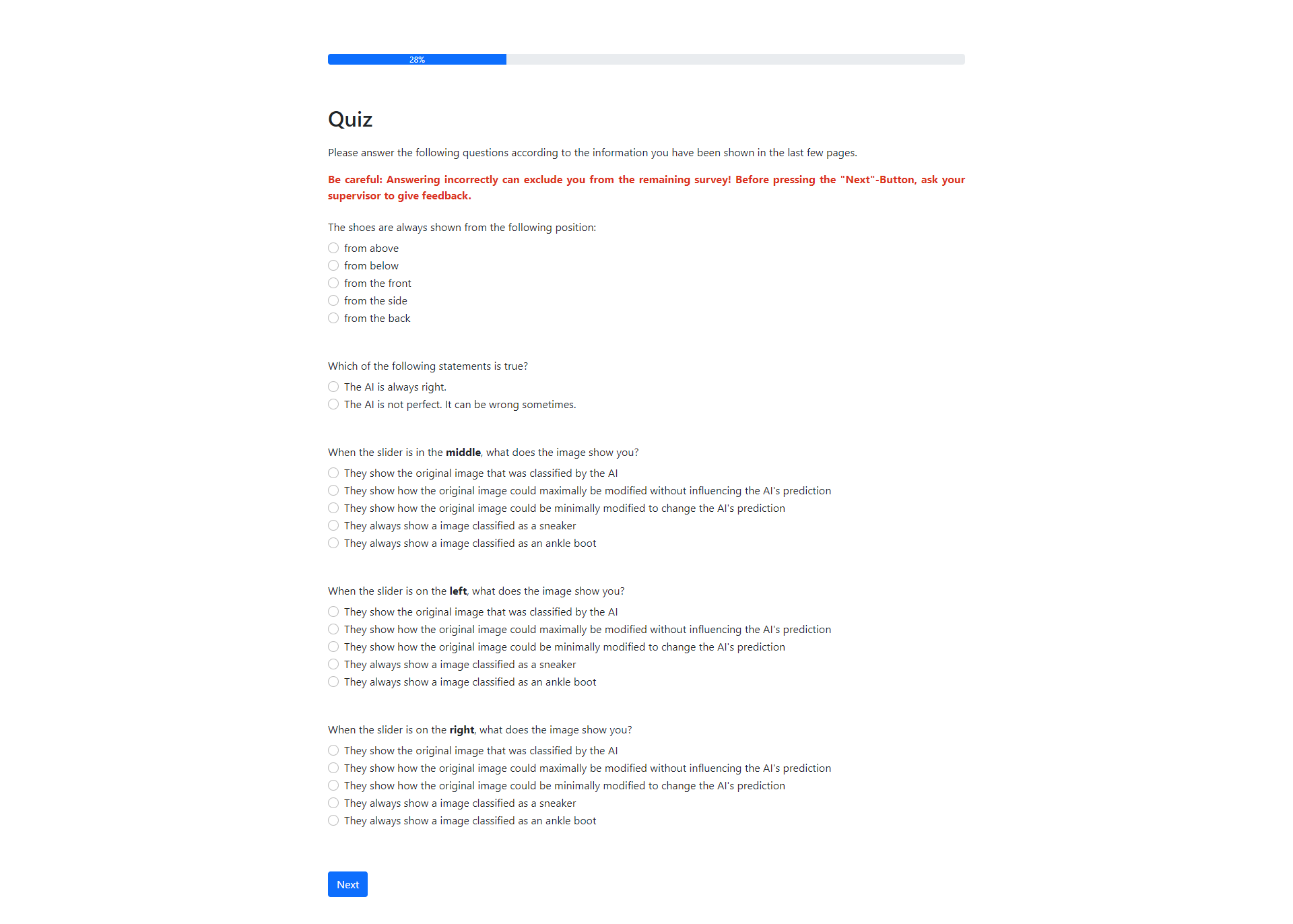}
    \caption{Screenshot of the user study, part 8.}
    \label{fig:study_screenshot_8}
\end{figure*}

\begin{figure*}
    \centering
    \includegraphics[width=\paperwidth]{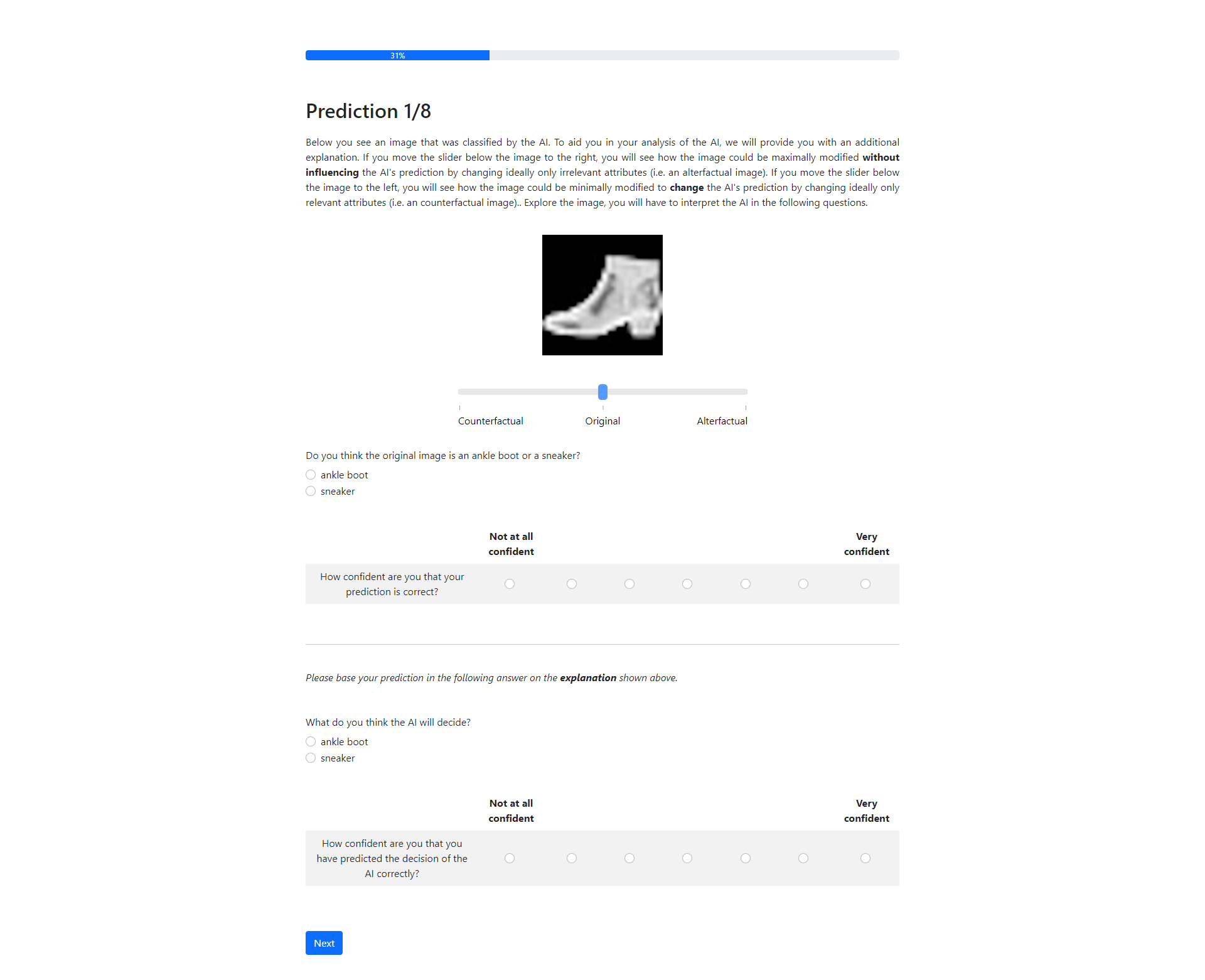}
    \caption{Screenshot of the user study, part 9.}
    \label{fig:study_screenshot_9}
\end{figure*}

\begin{figure*}
    \centering
    \includegraphics[width=\paperwidth]{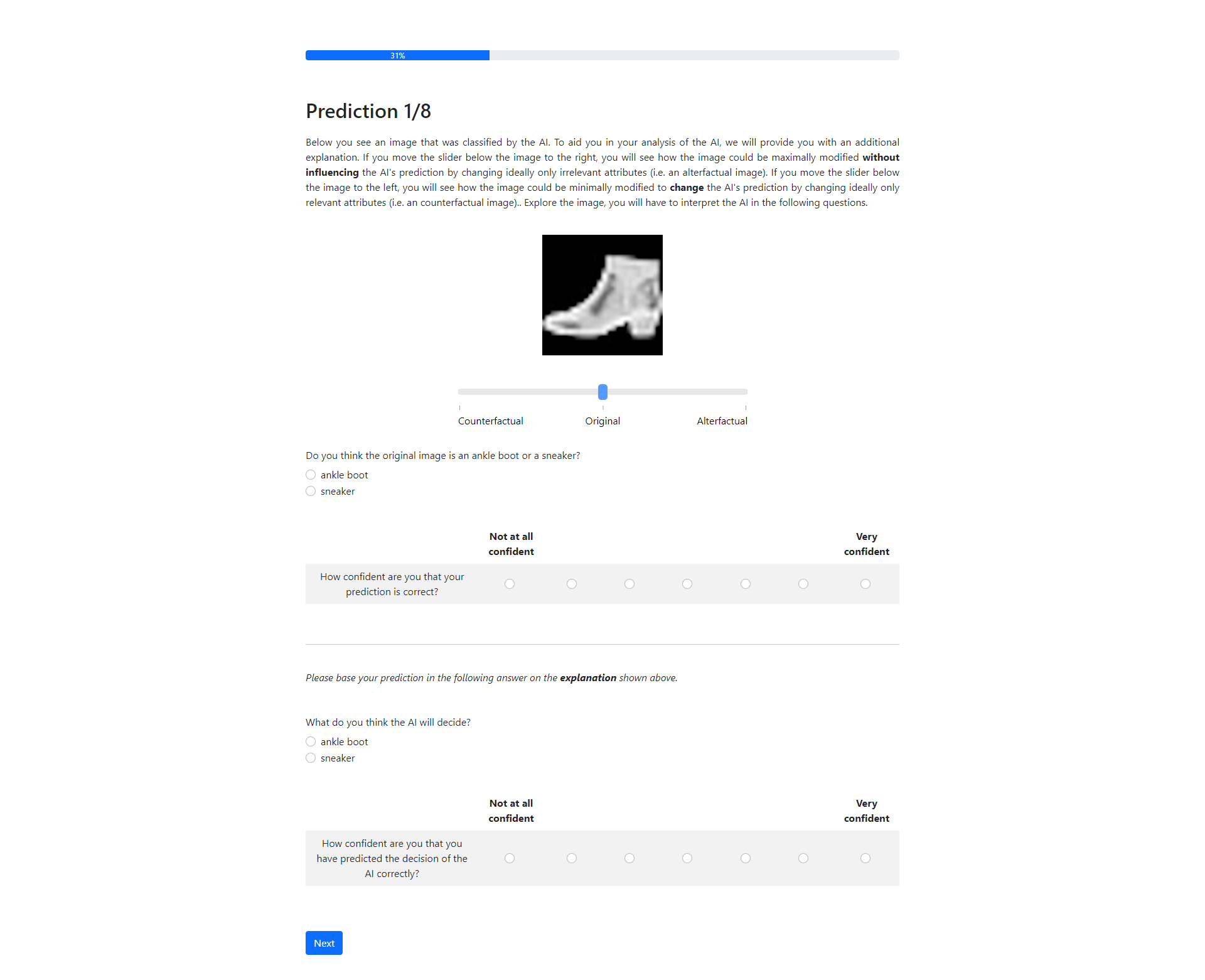}
    \caption{Screenshot of the user study, part 10.}
    \label{fig:study_screenshot_10}
\end{figure*}

\begin{figure*}
    \centering
    \includegraphics[width=\paperwidth]{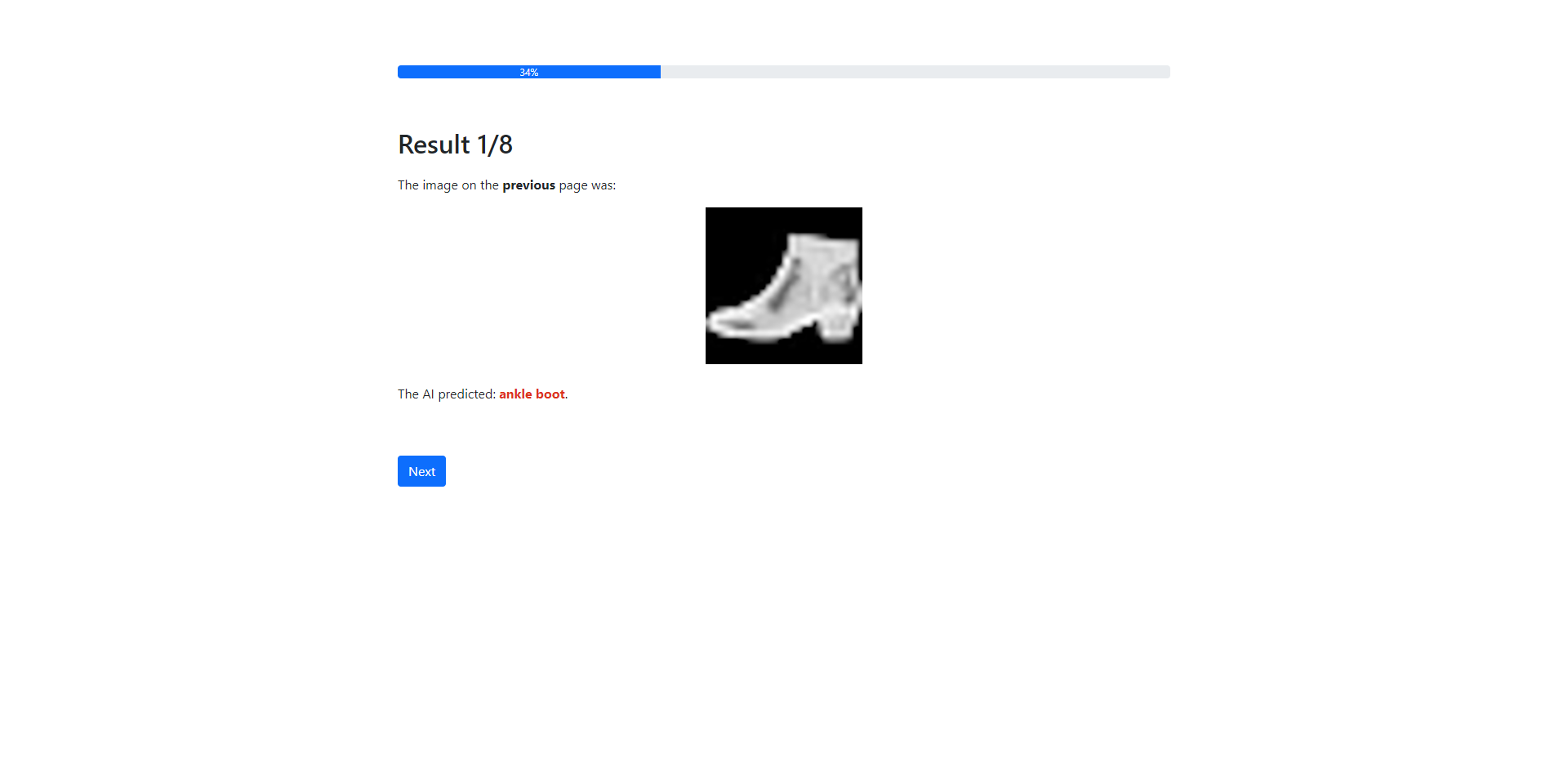}
    \caption{Screenshot of the user study, part 11.}
    \label{fig:study_screenshot_11}
\end{figure*}

\begin{figure*}
    \centering
    \includegraphics[width=\paperwidth]{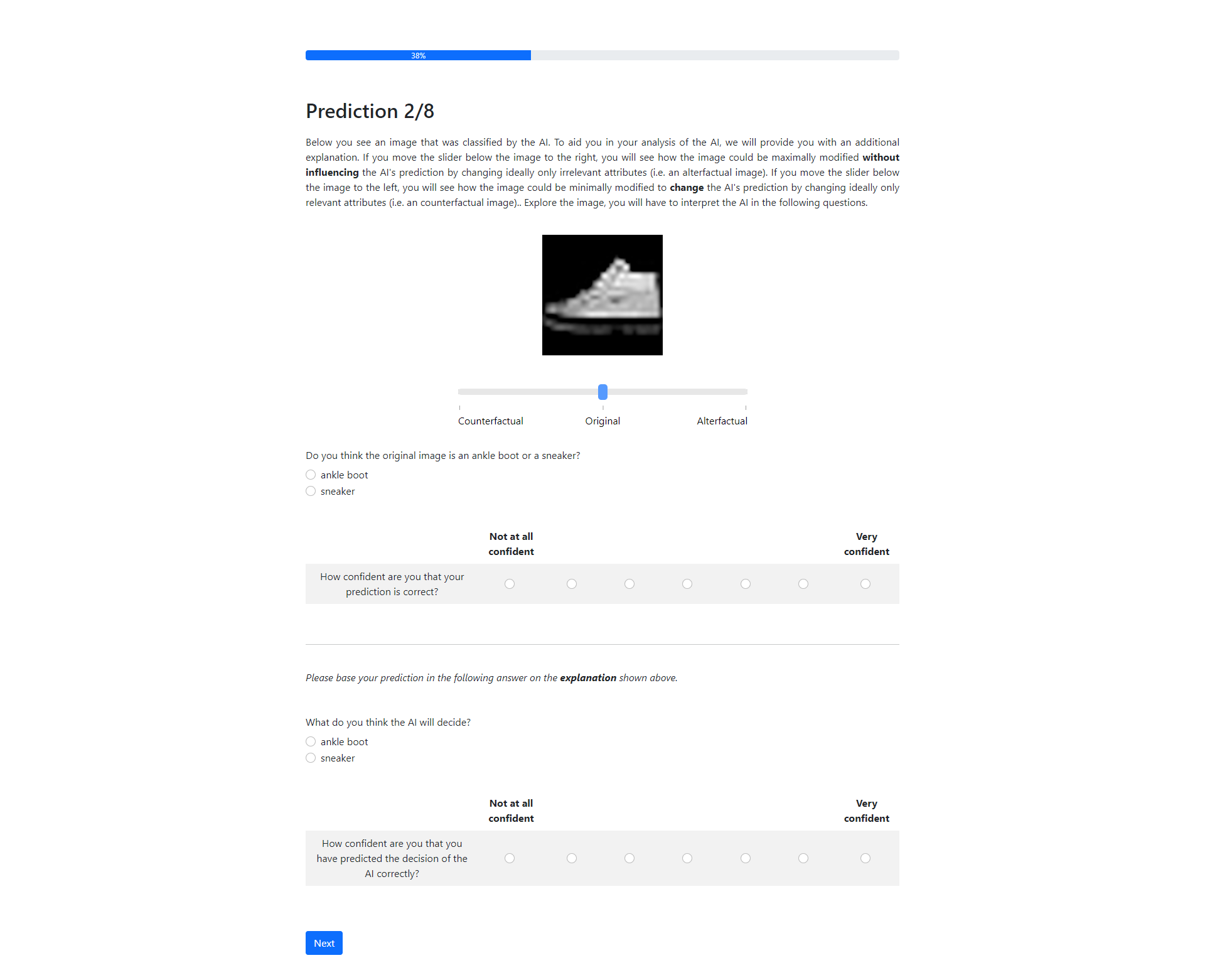}
    \caption{Screenshot of the user study, part 12.}
    \label{fig:study_screenshot_12}
\end{figure*}

\begin{figure*}
    \centering
    \includegraphics[width=\paperwidth]{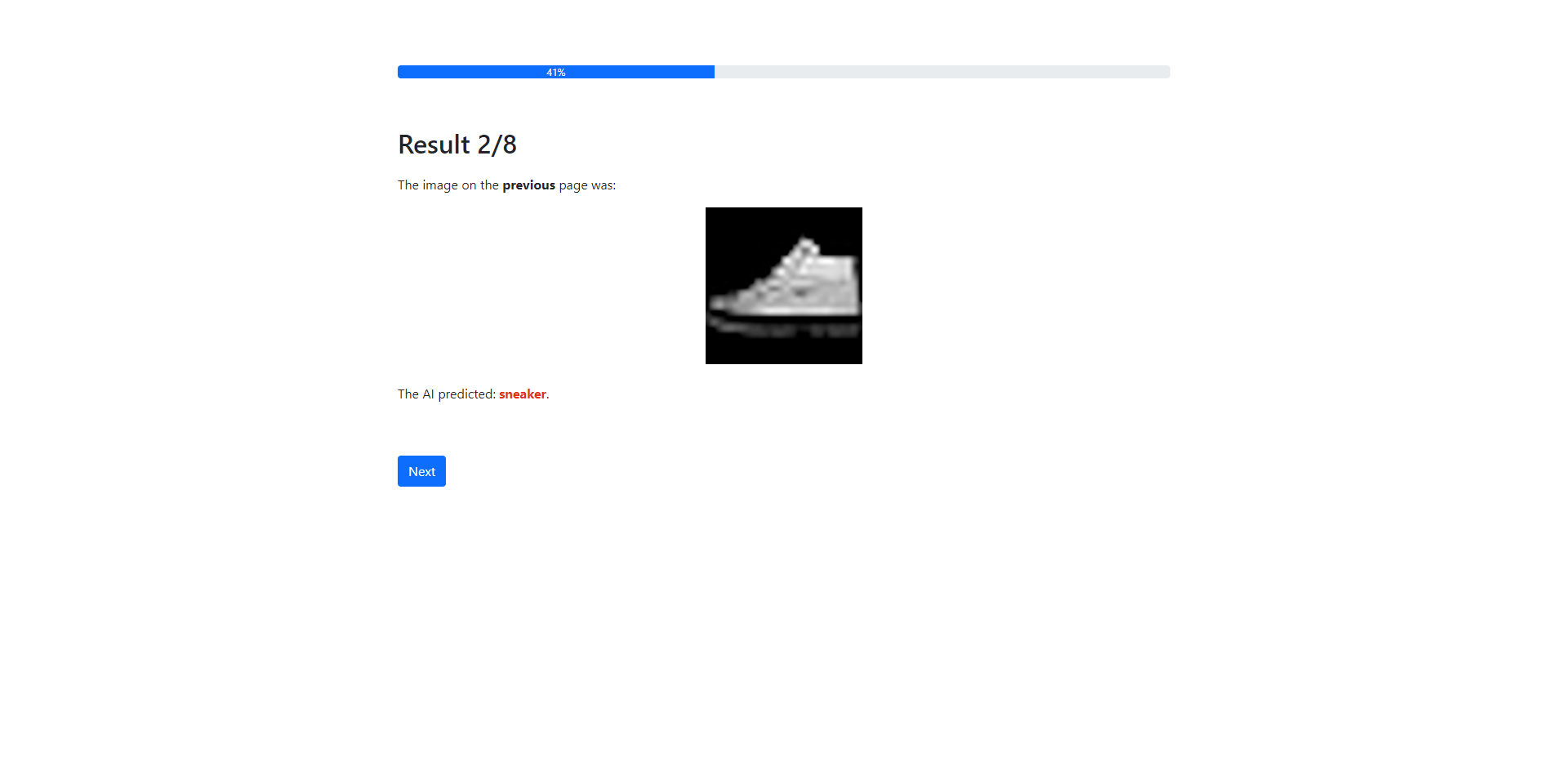}
    \caption{Screenshot of the user study, part 13.}
    \label{fig:study_screenshot_13}
\end{figure*}

\begin{figure*}
    \centering
    \includegraphics[width=\paperwidth]{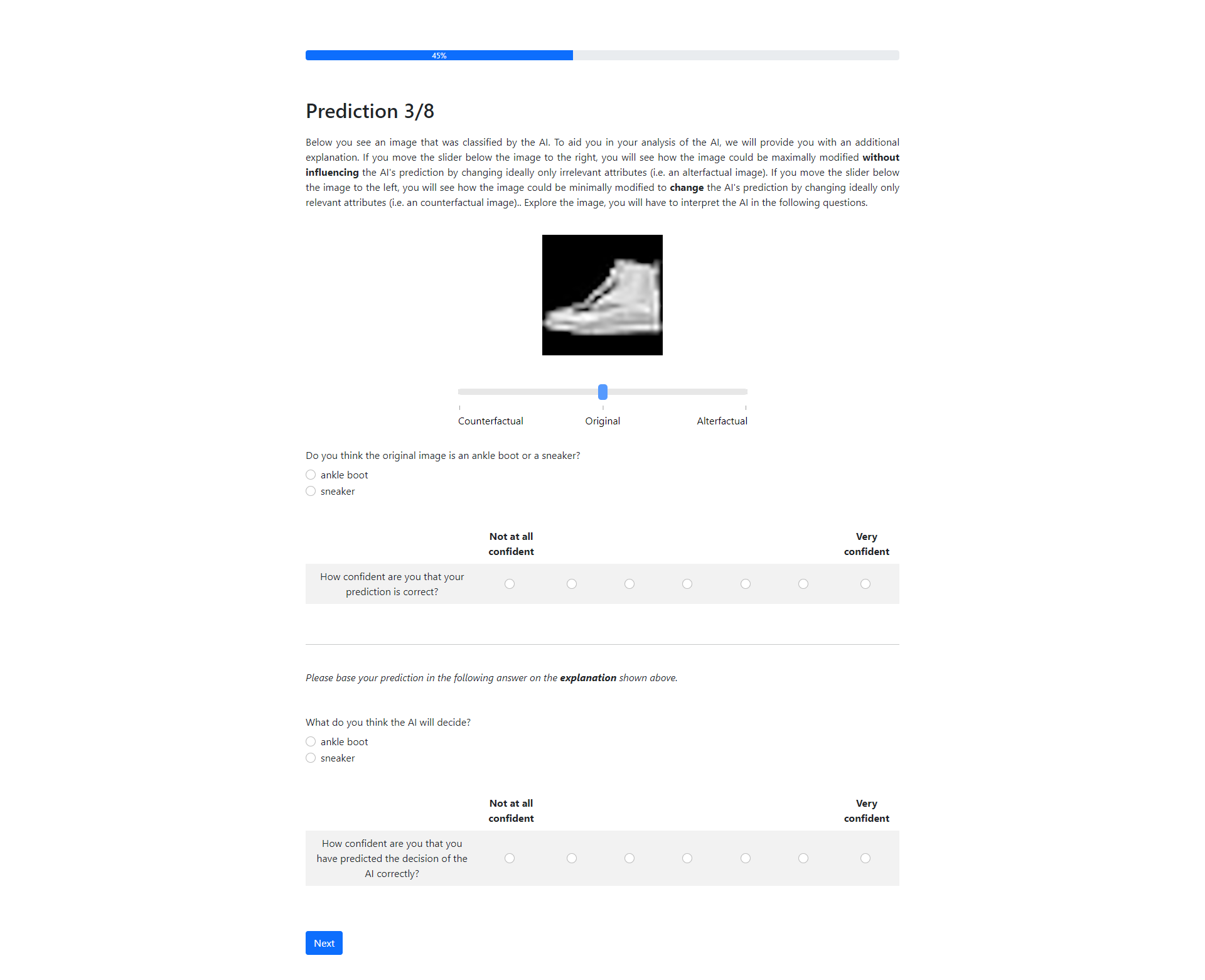}
    \caption{Screenshot of the user study, part 14.}
    \label{fig:study_screenshot_14}
\end{figure*}

\begin{figure*}
    \centering
    \includegraphics[width=\paperwidth]{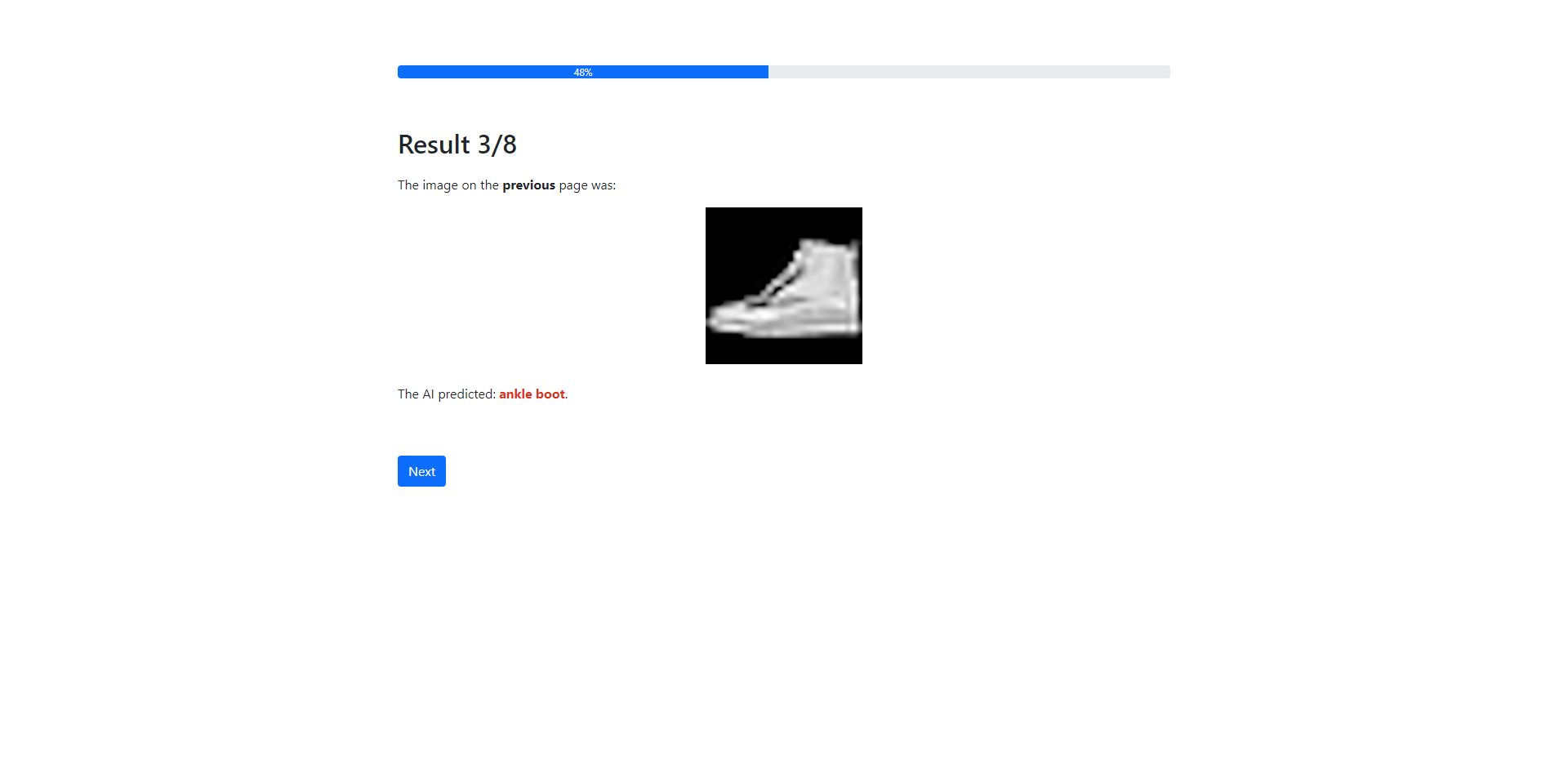}
    \caption{Screenshot of the user study, part 15.}
    \label{fig:study_screenshot_15}
\end{figure*}

\begin{figure*}
    \centering
    \includegraphics[width=\paperwidth]{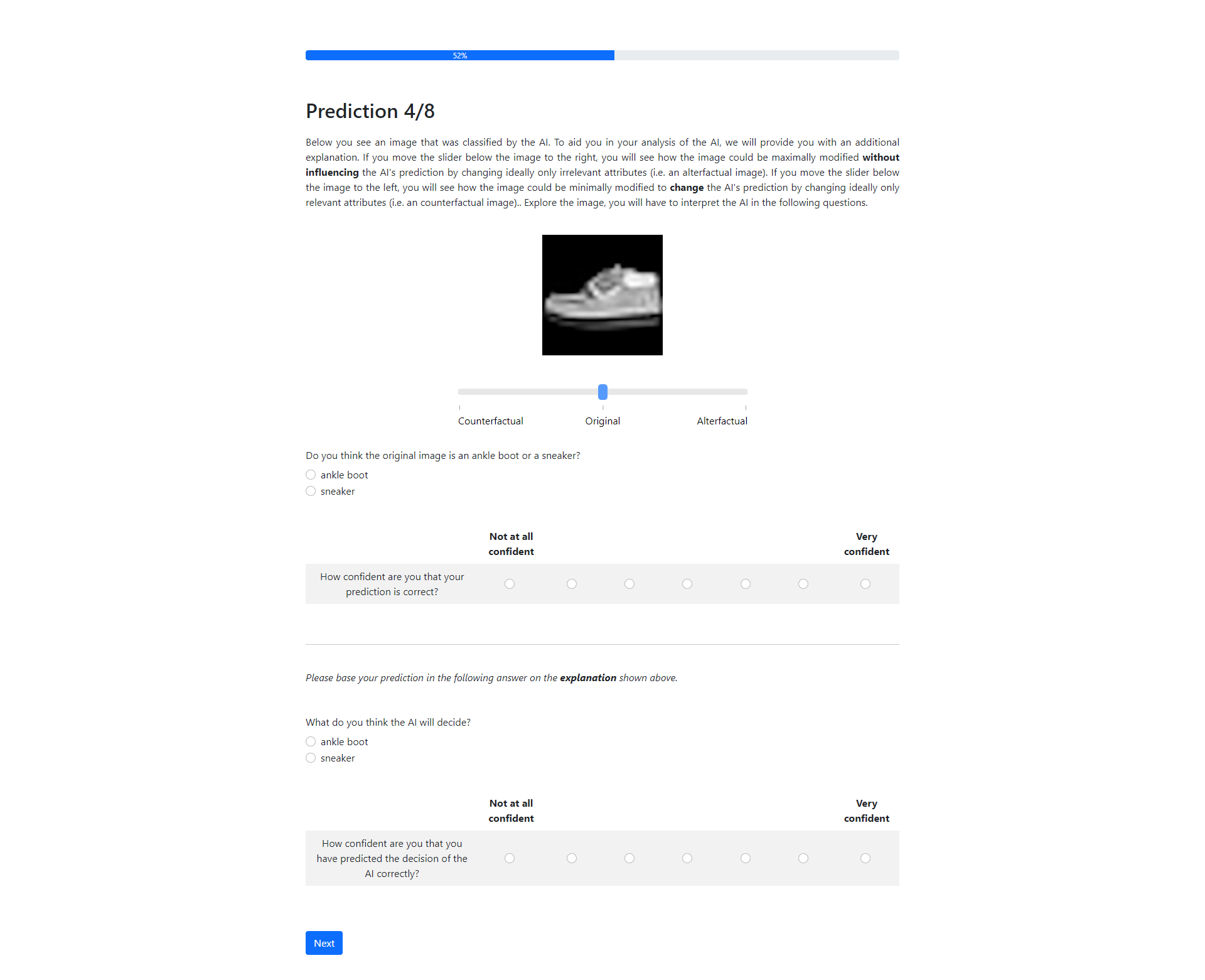}
    \caption{Screenshot of the user study, part 16.}
    \label{fig:study_screenshot_16}
\end{figure*}

\begin{figure*}
    \centering
    \includegraphics[width=\paperwidth]{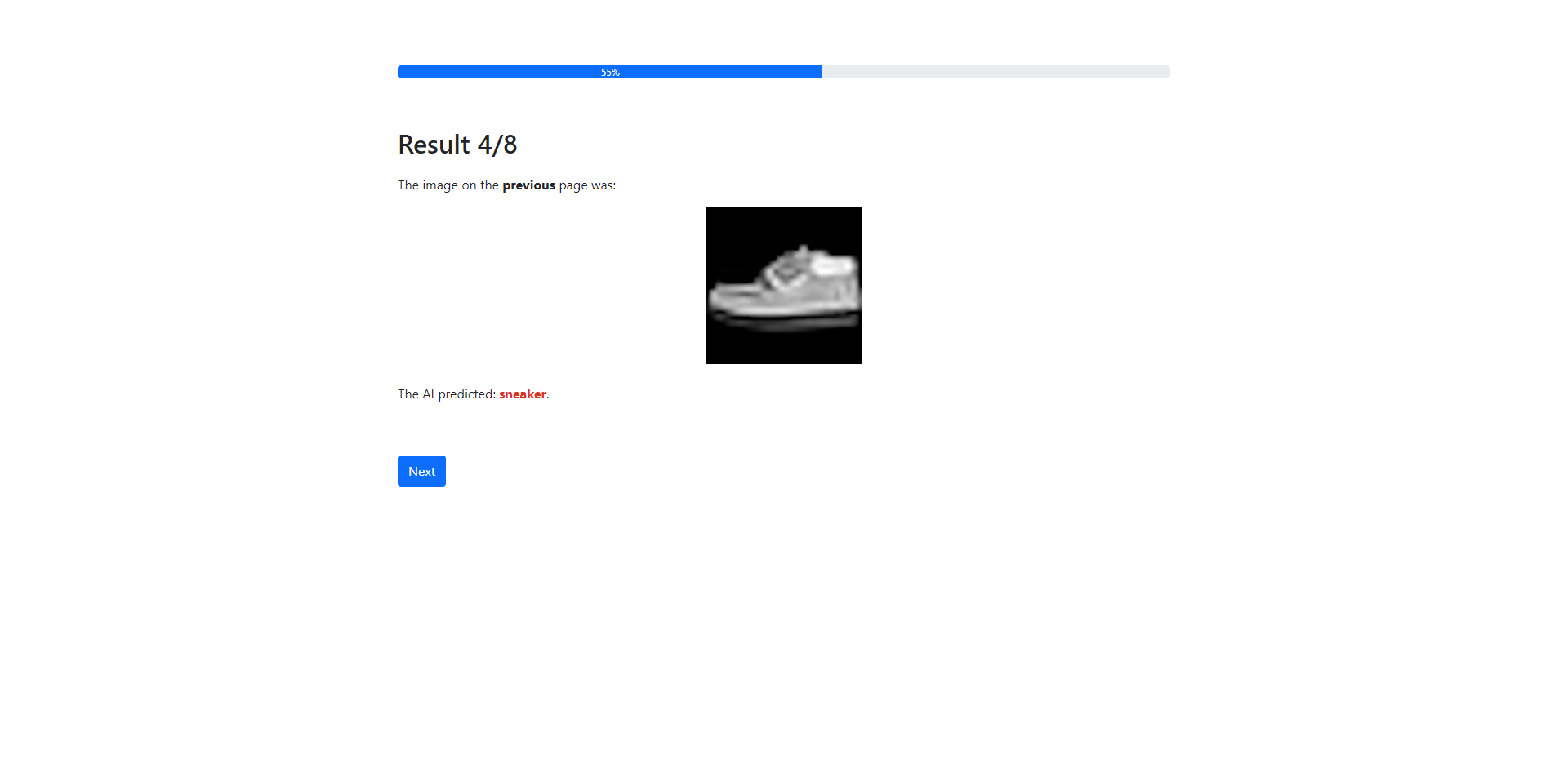}
    \caption{Screenshot of the user study, part 17.}
    \label{fig:study_screenshot_17}
\end{figure*}

\begin{figure*}
    \centering
    \includegraphics[width=\paperwidth]{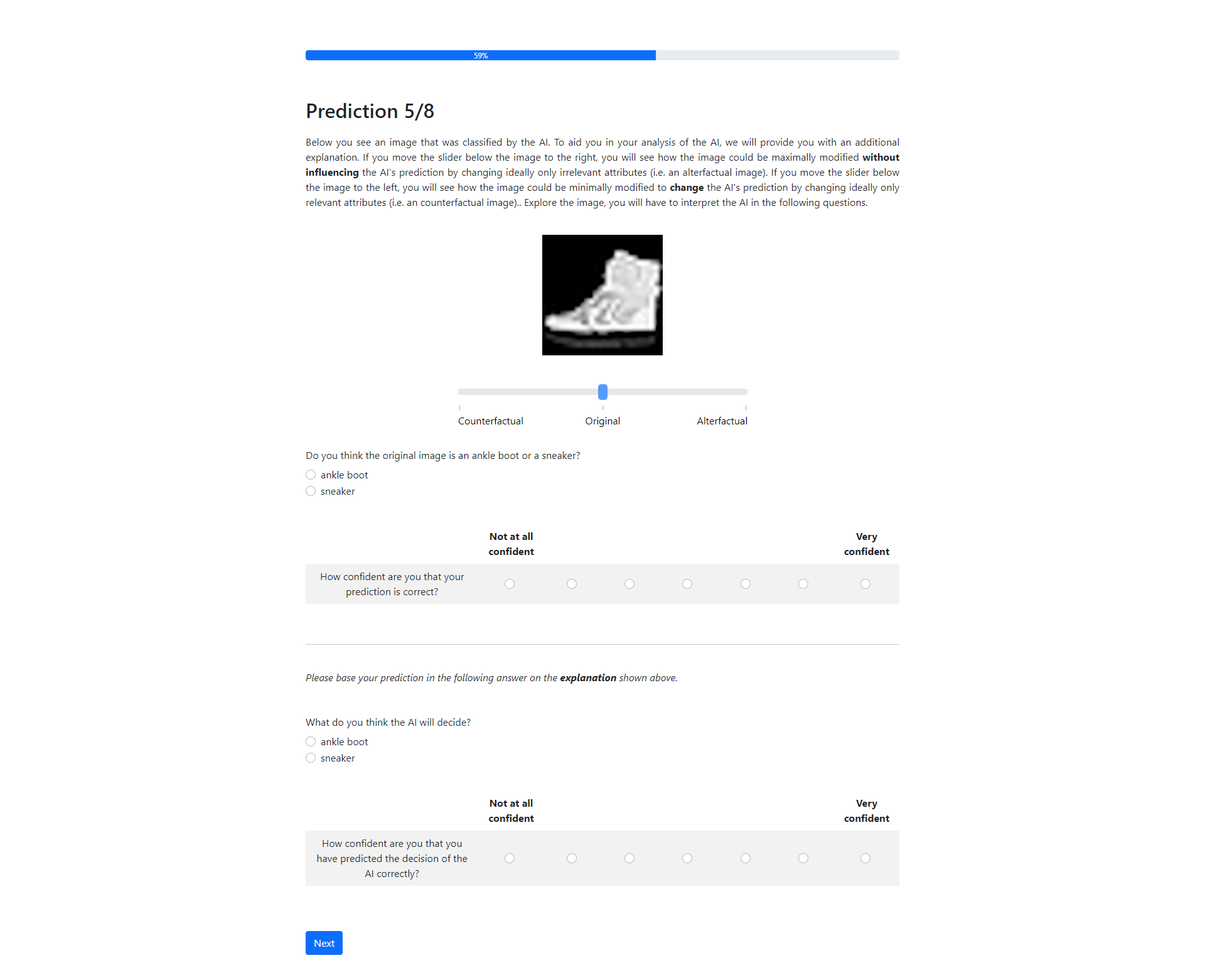}
    \caption{Screenshot of the user study, part 18.}
    \label{fig:study_screenshot_18}
\end{figure*}

\begin{figure*}
    \centering
    \includegraphics[width=\paperwidth]{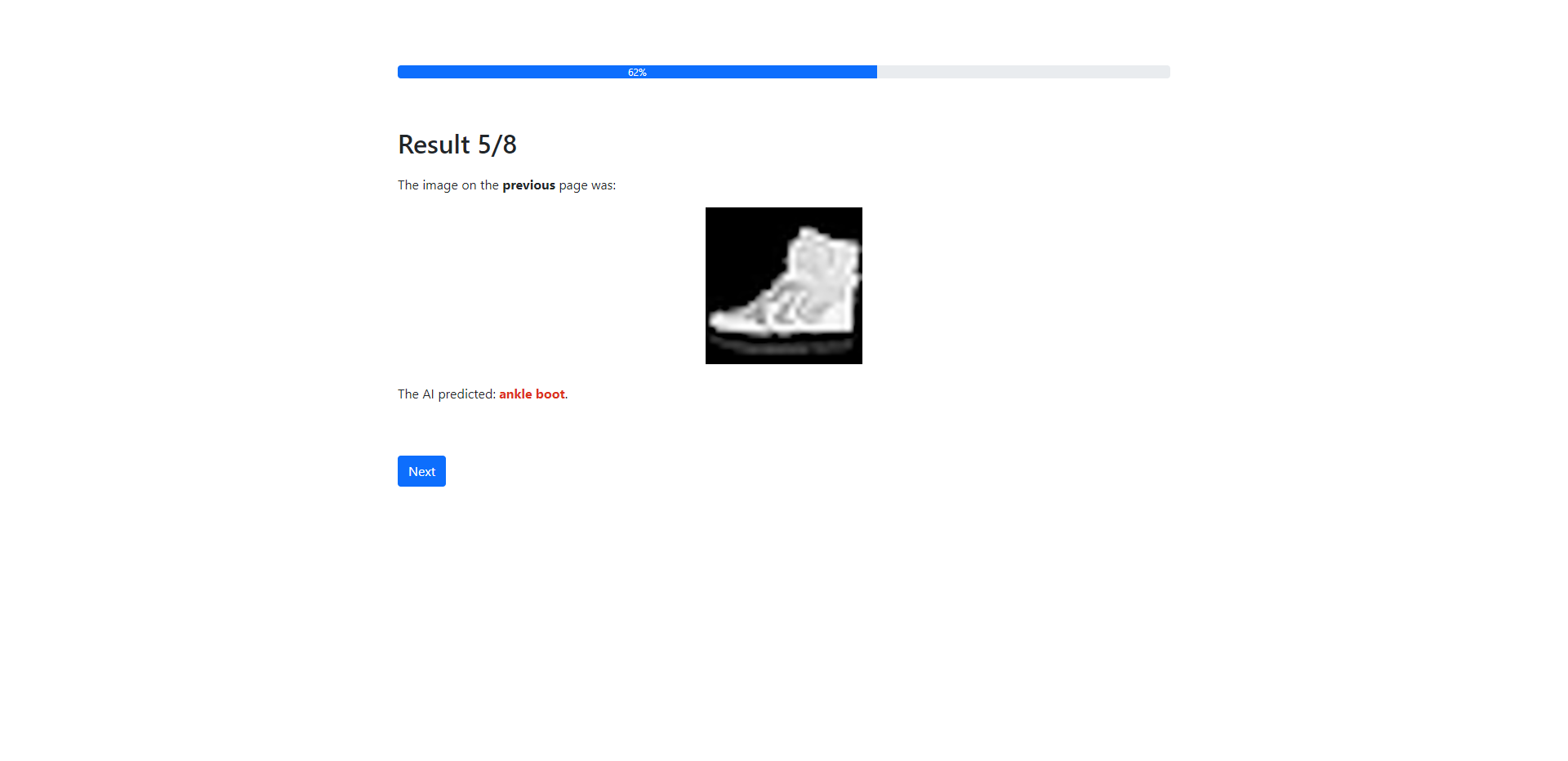}
    \caption{Screenshot of the user study, part 19.}
    \label{fig:study_screenshot_19}
\end{figure*}

\begin{figure*}
    \centering
    \includegraphics[width=\paperwidth]{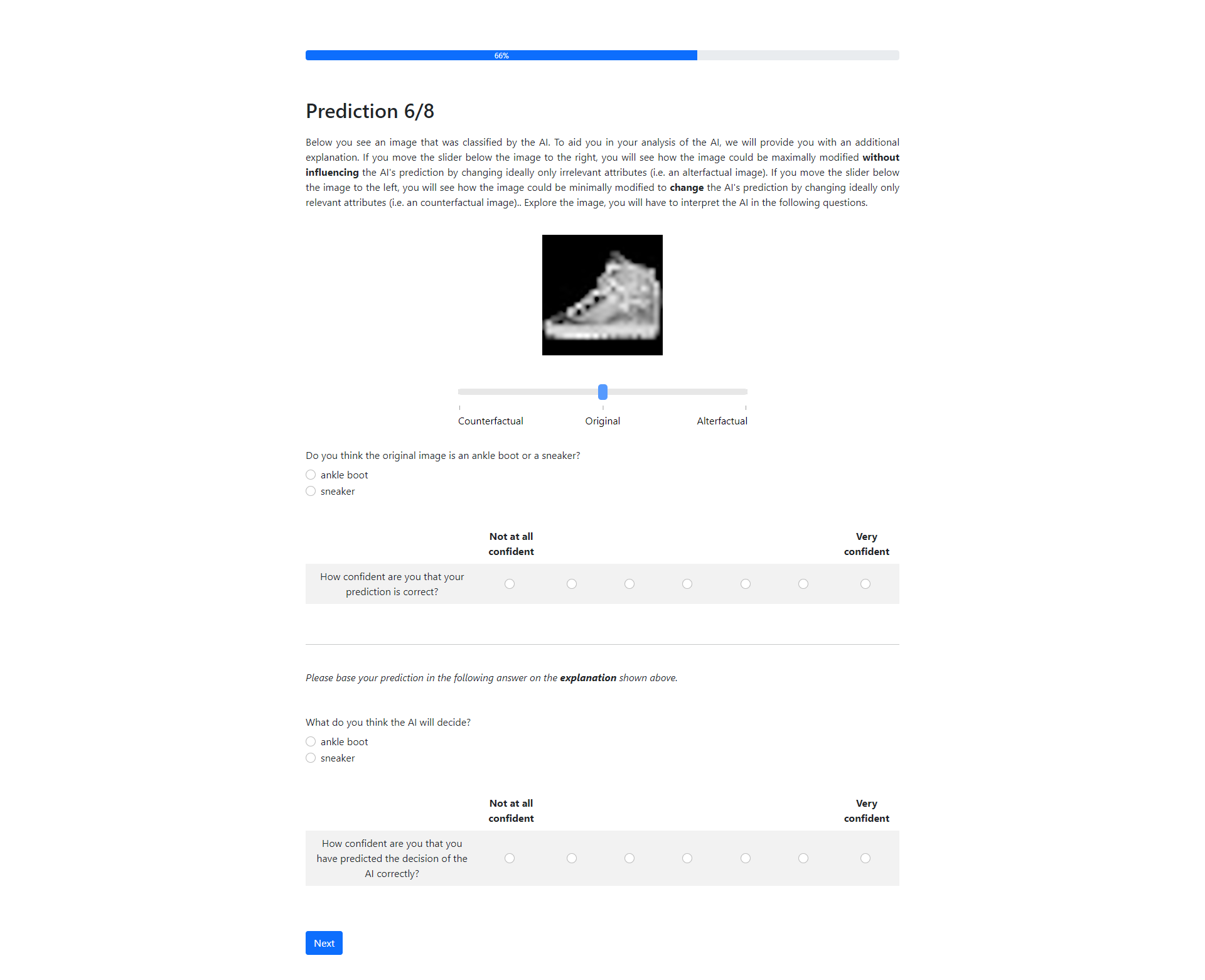}
    \caption{Screenshot of the user study, part 20.}
    \label{fig:study_screenshot_20}
\end{figure*}

\begin{figure*}
    \centering
    \includegraphics[width=\paperwidth]{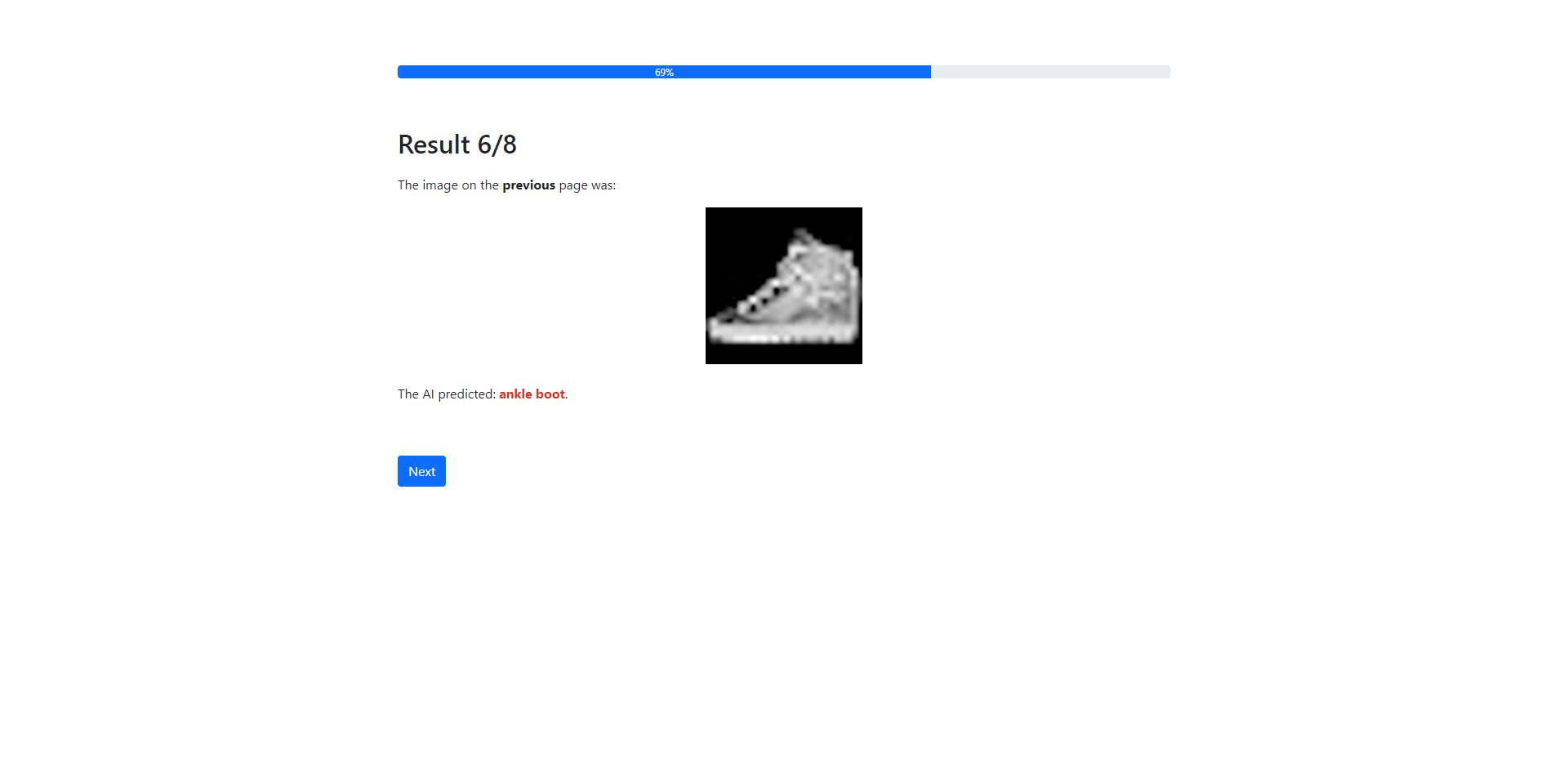}
    \caption{Screenshot of the user study, part 21.}
    \label{fig:study_screenshot_21}
\end{figure*}

\begin{figure*}
    \centering
    \includegraphics[width=\paperwidth]{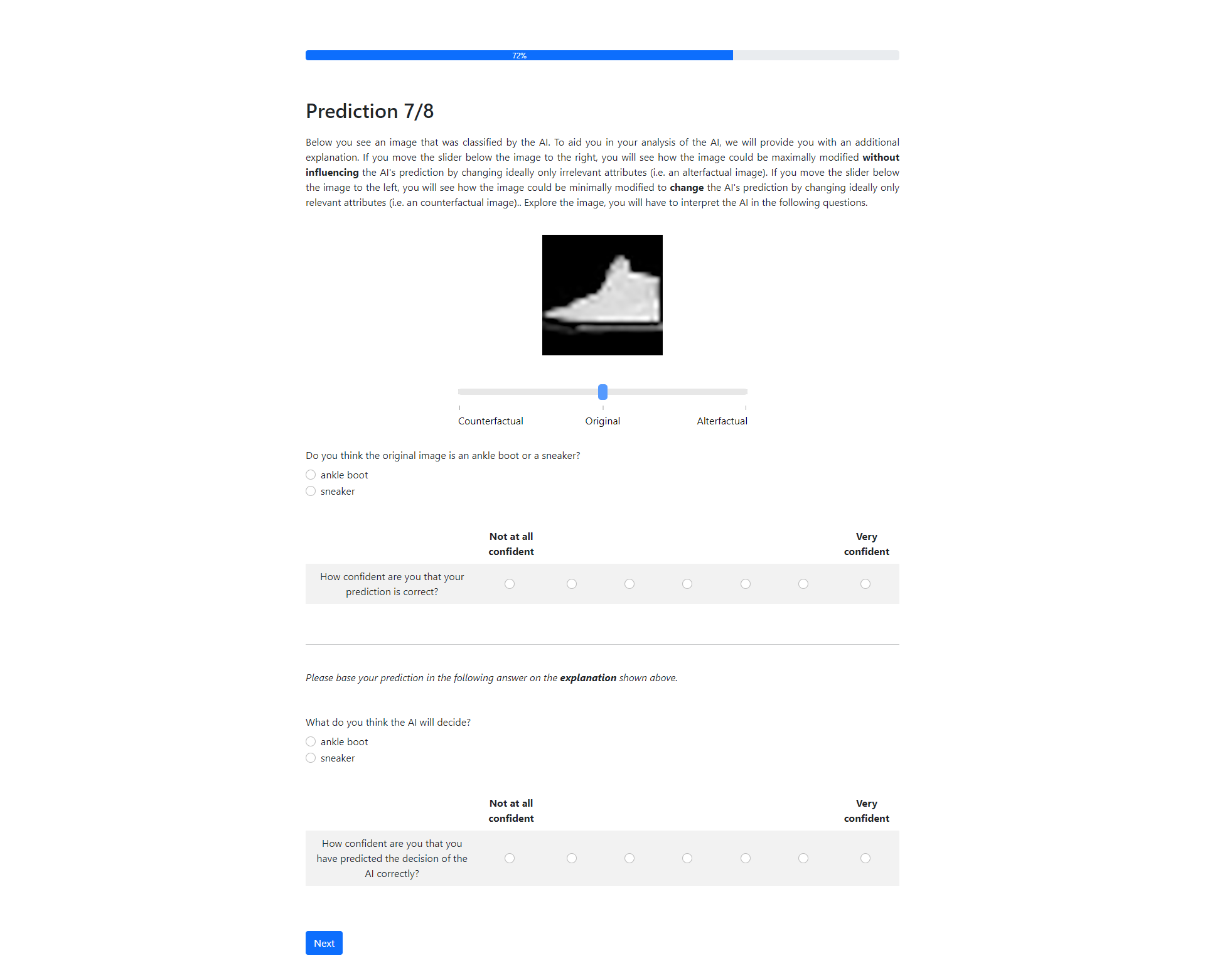}
    \caption{Screenshot of the user study, part 22.}
    \label{fig:study_screenshot_22}
\end{figure*}

\begin{figure*}
    \centering
    \includegraphics[width=\paperwidth]{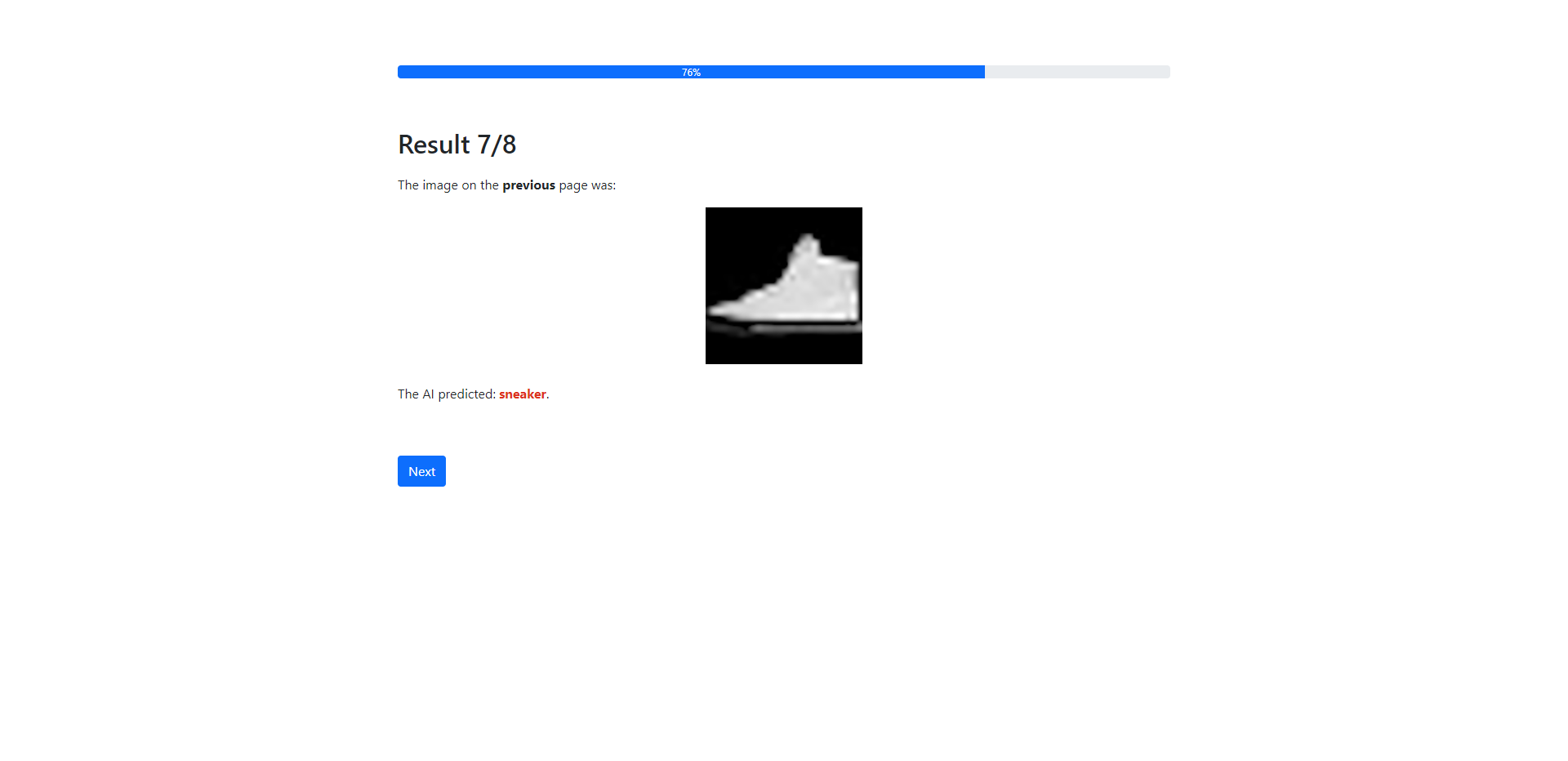}
    \caption{Screenshot of the user study, part 23.}
    \label{fig:study_screenshot_23}
\end{figure*}

\begin{figure*}
    \centering
    \includegraphics[width=\paperwidth]{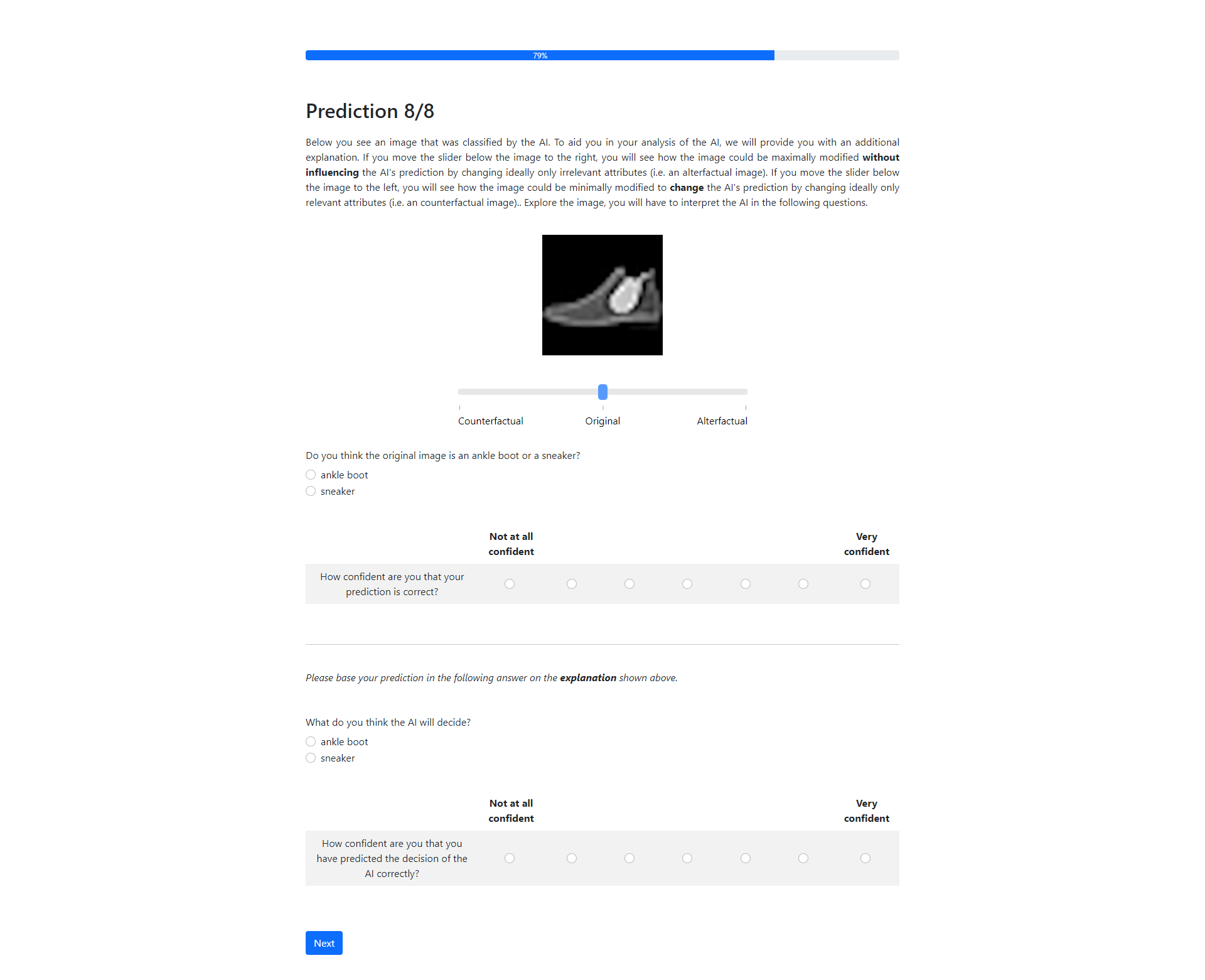}
    \caption{Screenshot of the user study, part 24.}
    \label{fig:study_screenshot_24}
\end{figure*}

\begin{figure*}
    \centering
    \includegraphics[width=\paperwidth]{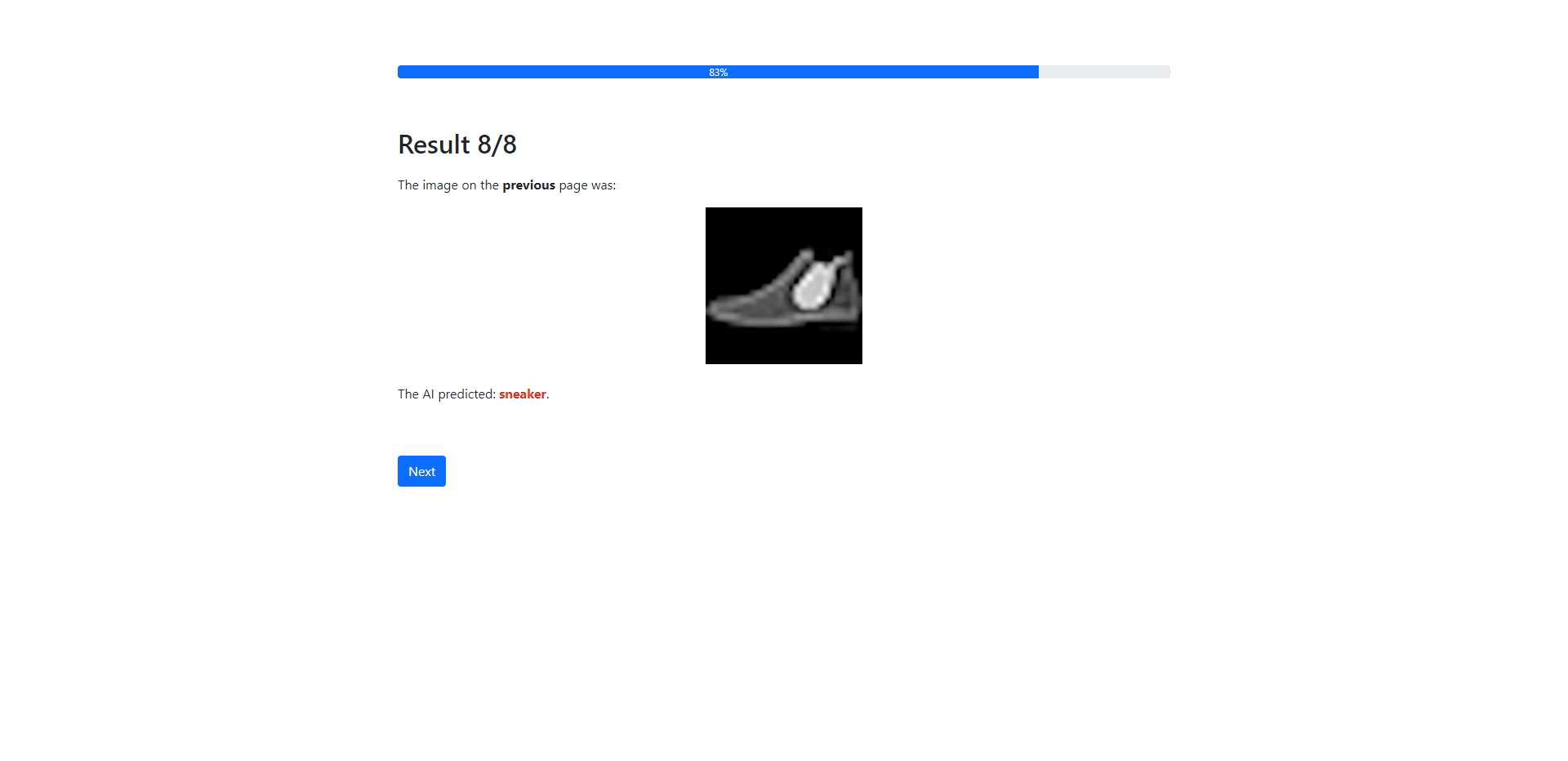}
    \caption{Screenshot of the user study, part 25.}
    \label{fig:study_screenshot_25}
\end{figure*}

\begin{figure*}
    \centering
    \includegraphics[width=\paperwidth]{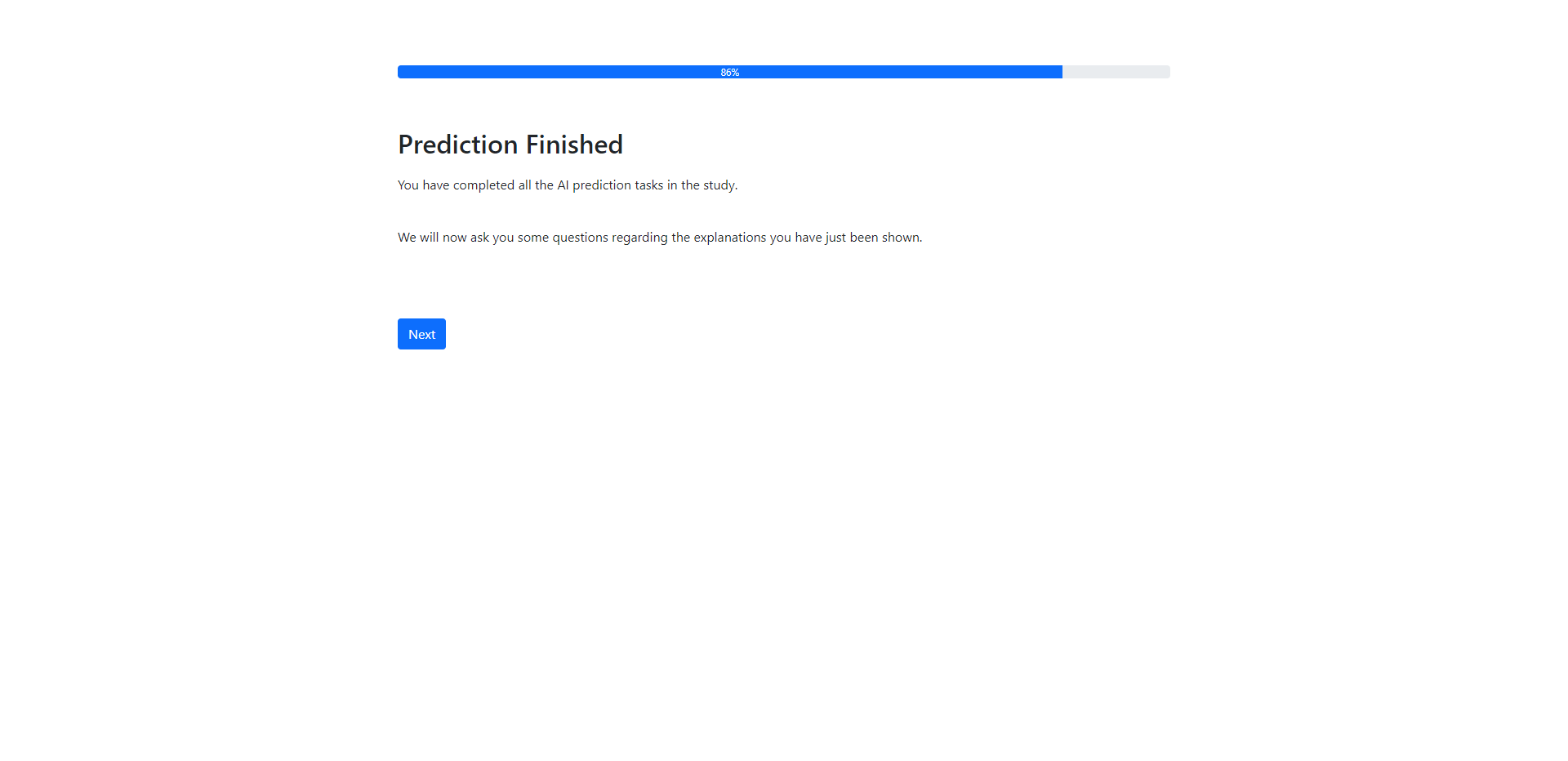}
    \caption{Screenshot of the user study, part 26.}
    \label{fig:study_screenshot_26}
\end{figure*}

\begin{figure*}
    \centering
    \includegraphics[width=0.8\paperwidth]{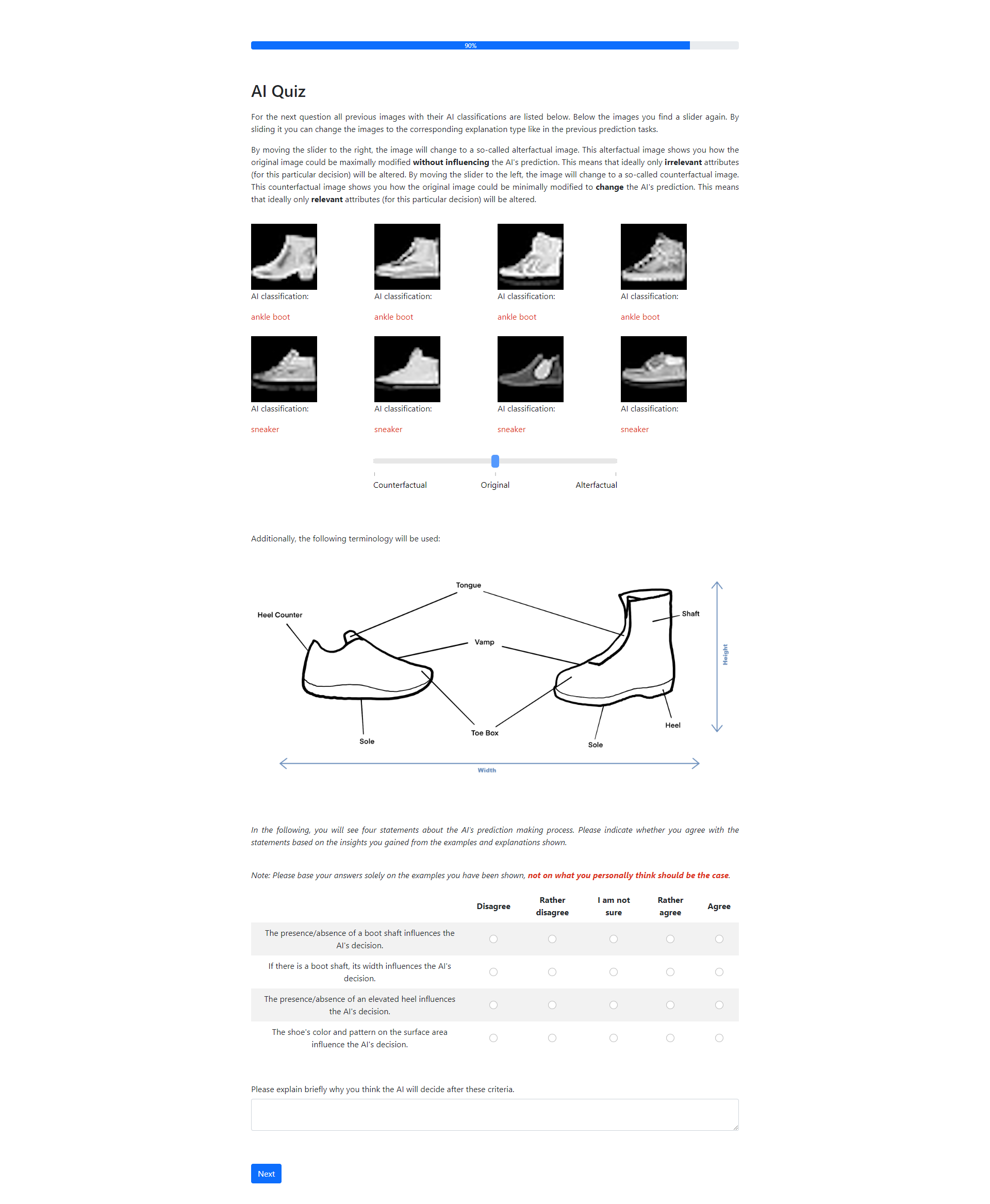}
    \caption{Screenshot of the user study, part 27.}
    \label{fig:study_screenshot_27}
\end{figure*}

\begin{figure*}
    \centering
    \includegraphics[width=\paperwidth]{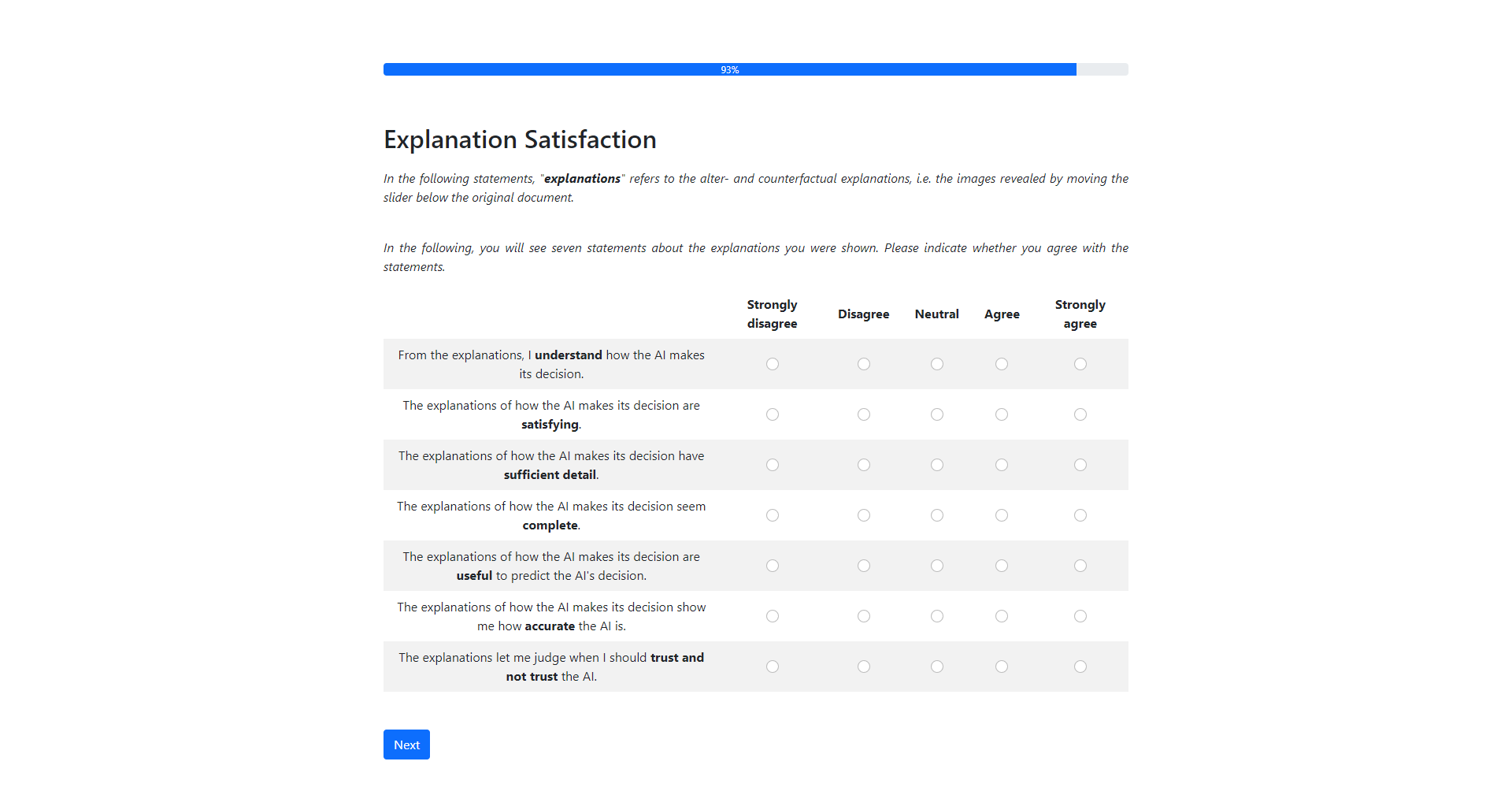}
    \caption{Screenshot of the user study, part 28.}
    \label{fig:study_screenshot_28}
\end{figure*}

\begin{figure*}
    \centering
    \includegraphics[width=\paperwidth]{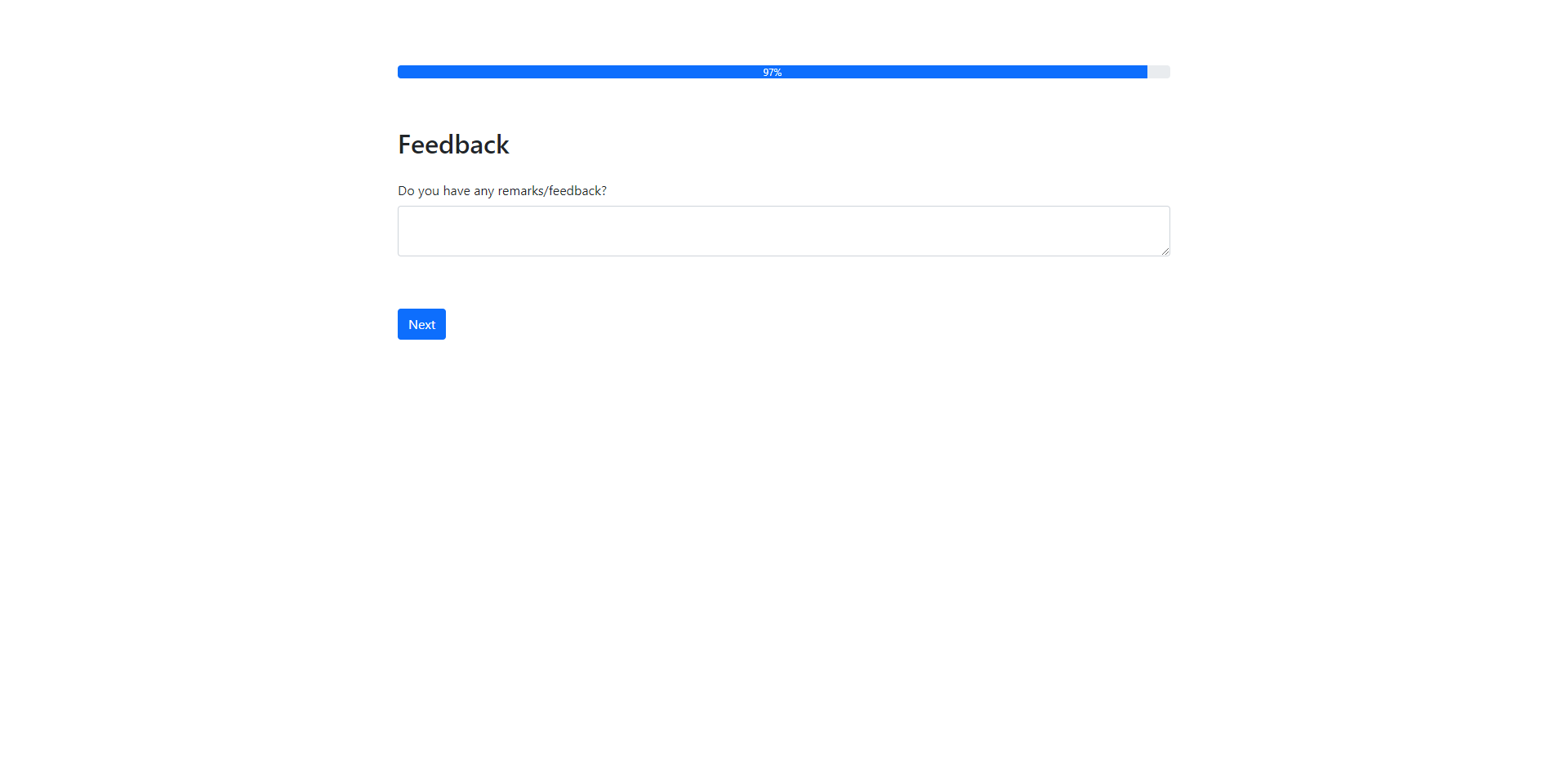}
    \caption{Screenshot of the user study, part 29.}
    \label{fig:study_screenshot_29}
\end{figure*}

\begin{figure*}
    \centering
    \includegraphics[width=\paperwidth]{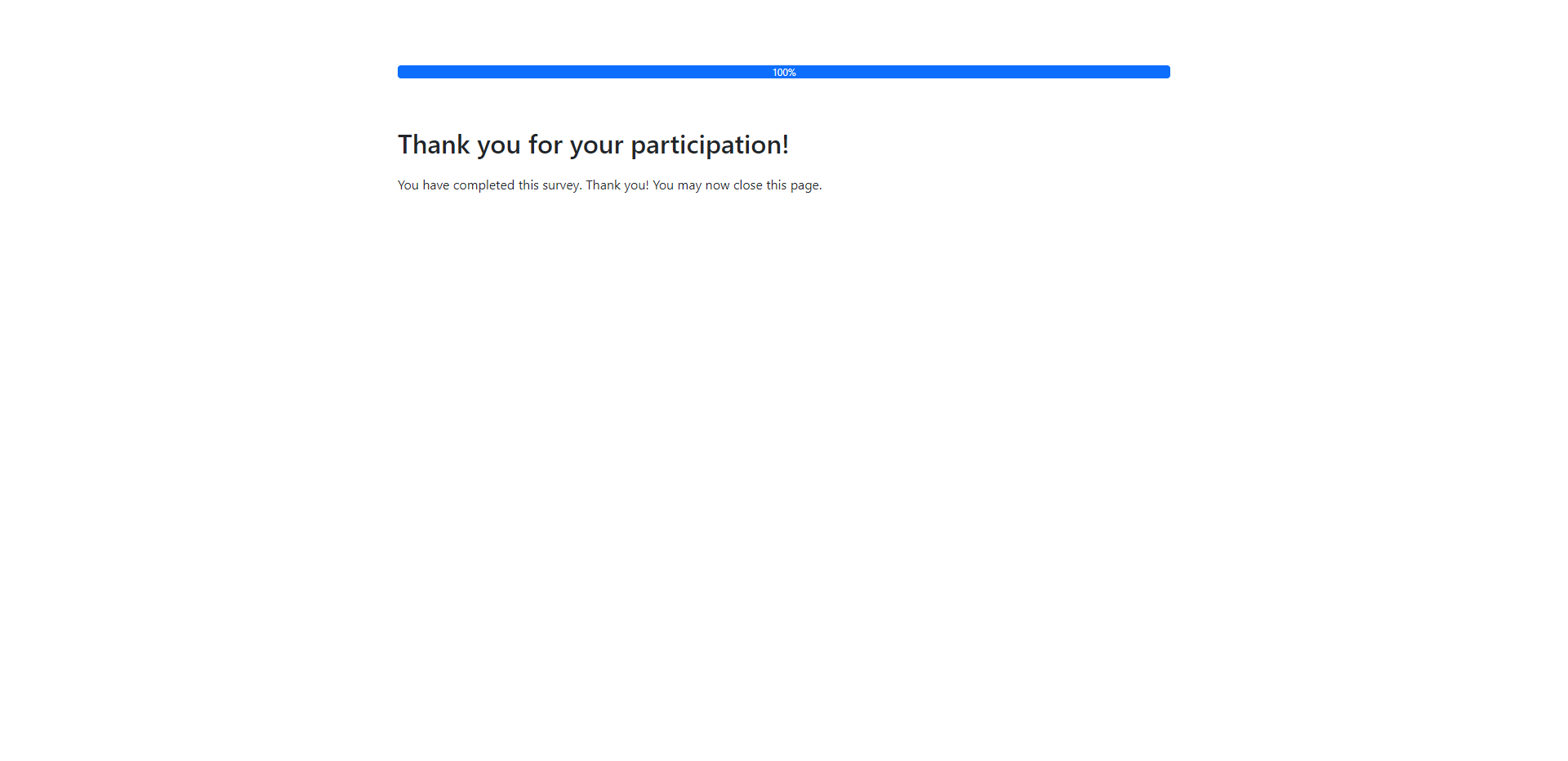}
    \caption{Screenshot of the user study, part 30.}
    \label{fig:study_screenshot_30}
\end{figure*}

Screenshots of the user study are shown in Figures \ref{fig:study_screenshot_1} to \ref{fig:study_screenshot_30}. The \Combination{} condition is shown.

\begin{table*}
    \small
	\centering
	\rowcolors{2}{white}{lightgray}
	%\rotatebox{90}{
	\begin{tabular}{c l c c c c c c} 
		Layer   & Description       & \# Filter     & Size  & Stride     & Dropout   & BatchNorm & Activation\\
		\hline
        1       & Conv2D            & 64            & 4x4   & 2         & -         & no         & LeakyReLU (0.2)\\
        2       & Conv2D            & 128           & 4x4   & 2         & -         & yes        & LeakyReLU (0.2)\\
        3       & Conv2D            & 256           & 4x4   & 2         & -         & yes        & LeakyReLU (0.2)\\
        4       & Conv2D            & 512           & 4x4   & 2         & -         & yes        & LeakyReLU (0.2)\\
        5       & Conv2D            & 512           & 4x4   & 2         & -         & yes        & LeakyReLU (0.2)\\
        6       & Conv2D            & 512           & 4x4   & 2         & -         & yes        & LeakyReLU (0.2)\\
        7       & Conv2D            & 512           & 4x4   & 2         & -         & no         & ReLU\\
        8       & Conv2DTranspose   & 512           & 4x4   & 2         & 0.5       & yes        & ReLU\\
        9       & Conv2DTranspose   & 512           & 4x4   & 2         & 0.5       & yes        & ReLU\\
        10      & Conv2DTranspose   & 512           & 4x4   & 2         & 0.5       & yes        & ReLU\\
        11      & Conv2DTranspose   & 256           & 4x4   & 2         & -         & yes        & ReLU\\
        12      & Conv2DTranspose   & 128           & 4x4   & 2         & -         & yes        & ReLU\\
        13      & Conv2DTranspose   & 64            & 4x4   & 2         & -         & yes        & ReLU\\
        14      & Conv2DTranspose   & 1             & 4x4   & 2         & -         & no         & Tanh\\
	\end{tabular}
	%}
	\caption[]{Generator Architecture used in our evaluation scenario. The architecture is adapted from \citet{wu2019privacy}. Where BatchNorm, Dropout, or Activation function occurred together, the order applied was BatchNorm - Dropout - Activation.} 
	\label{table:gen}
\end{table*}

\begin{table*} 
	\centering
	\rowcolors{2}{white}{lightgray}
	%\rotatebox{90}{
	\begin{tabular}{c l c c c c c c} 
		Layer   & Description       & \# Filter     & Size      & Stride       & BatchNorm & Activation\\
		\hline
        0a      & Embedding         & -             & 8x8       & -                  & no        & -\\
        0b      & Upsample          & -             & 128x128   & -                  & no        & -\\
        1       & Conv2D            & 64            & 4x4       & 2                  & no        & LeakyReLU (0.2)\\
        2       & Conv2D            & 128           & 4x4       & 2                  & yes       & LeakyReLU (0.2)\\
        3       & Conv2D            & 256           & 4x4       & 2                  & yes       & LeakyReLU (0.2)\\
        4       & Conv2D            & 1             & 4x4       & 2                  & no        & Sigmoid\\
	\end{tabular}
	%}
	\caption[]{Discriminator Architecture used in our evaluation scenario. Where BatchNorm and Activation function occurred together, BatchNorm preceded the activation function. The first two layers, marked as `0a' and `0b' were used to upsample the label information to the size of the input image. The label and image were passed together to layer 1. The architecture is adapted from \citet{wu2019privacy}.}
	\label{table:d1}
\end{table*}

\begin{table*}
    \small
	\centering
	\rowcolors{2}{white}{lightgray}
	\begin{tabular}{c | c} 
        Batch Size     & 1 \\
        Epochs         & 14 \\
        Learning Rate Generator  & 1e-4 \\
        Learning Rate Discriminator  & 1e-4 \\
        Optimizer      & Adam  \\
	\end{tabular}
	\caption[]{The setting used to train the GAN.}
	\label{table:gan_configs}
\end{table*}

\begin{table*}
    \small
	\centering
	\rowcolors{2}{white}{lightgray}
	\begin{tabular}{c | c} 
        C (Regularisation)     & 10 \\
        Kernel         & linear \\
        Iterations  & 5000 \\
	\end{tabular}
	\caption[]{The setting used to train the SVM.}
	\label{table:svm_configs}
\end{table*}

\begin{table*}[]    
	\centering
	\rowcolors{2}{white}{lightgray}
 %\rotatebox{90}{
	\begin{tabular}{c l c c c c c c} 
		Layer   & Description   & \# Filter     & Size  & Stride       & BatchNorm & Activation\\
		\hline
        1       & Conv2D        & 32            & 3x3   & 1                 & yes       & ReLU\\
        2       & Conv2D        & 32            & 3x3   & 1                  & yes       & ReLU\\
        3       & MaxPool2D     & -             & 2x2   & 2                 & no        & - \\
        4       & Conv2D        & 64            & 3x3   & 1                  & yes       & ReLU\\
        5       & Conv2D        & 64            & 3x3   & 1                  & yes       & ReLU\\
        6       & GAP           & -             & -     & -                  & no        & -\\
        7       & Dense         & -             & 2     & -                  & no        & Softmax\\
	\end{tabular}
 %}
	\caption[]{Classifier architecture used to train the classifier for the MNIST-Fashion dataset (classes \emph{Sneaker} and \emph{Ankle Boot}). Where BatchNorm and Activation function occurred together, BatchNorm preceded the activation function.}
	\label{table:cl1}
\end{table*}

\begin{table*}
    \small
	\centering
	\rowcolors{2}{white}{lightgray}
	\begin{tabular}{c | c} 
        Batch Size     & 32 \\
        Epochs         & 40 \\
        Learning Rate  & 1e-3 \\
        Optimizer      & Adam \\
        Loss Function  & Binary Cross Entropy \\
	\end{tabular}
	\caption[]{The setting used to train the Fashion-MNIST classifier.}
	\label{table:classifier_configs}
\end{table*}

\begin{table*}
\small
\centering
\rowcolors{2}{white}{lightgray}
 %\rotatebox{90}{
\begin{tabular}{ccccc} % Use "c" for centered columns

 & Control & Counterfactual & Alterfactual & Combination \\
\midrule
Mean age & 22.0 & 22.5 & 21.6 & 22.6 \\
Percentage of female participants & 52 & 58 & 47 & 58 \\
Highest level of Education & 2.27 & 2.42 & 2.19 & 2.21 \\
Mean AI Attitude & 3.50 & 3.82 & 3.84 & 3.75 \\
\bottomrule
\end{tabular}
%}
\caption{Demographic data across conditions. The highest level of education was measured as follows: 1 - No education, 2 - High school graduation, 3 - Vocational training, 3 - Bachelor, 4 - Master, 5 - Doctor. The Attitude towards AI is measured on 5-Point Likert scale from ``Extremely negative'' to ``Extremely positive''.}
\label{tab:demographics_per_condition}
\end{table*}

{}

{}

{}

{}

{}

{}

{}

{}

{}

{}

{}

{}

{}

{}

{}

{}

{}

{}

{}

{}

{}

{}

{}

{}

{}

{}

{}

{}

{}

{}

\end{document}